\documentclass[twoside]{article}

\usepackage[preprint]{aistats2027}
\usepackage[round]{natbib}
\renewcommand{\bibname}{References}
\renewcommand{\bibsection}{\subsubsection*{\bibname}}

\usepackage{amsmath,amsfonts,bm}

\def\eqref#1{equation~\ref{#1}}

\def\1{\bm{1}}

\DeclareMathAlphabet{\mathsfit}{\encodingdefault}{\sfdefault}{m}{sl}
\SetMathAlphabet{\mathsfit}{bold}{\encodingdefault}{\sfdefault}{bx}{n}

\usepackage[utf8]{inputenc} 
\usepackage[T1]{fontenc}    
\usepackage{lmodern}        
\usepackage{url}            
\usepackage{booktabs}       
\usepackage{multirow}
\usepackage{amsfonts}       
\usepackage{nicefrac}       
\usepackage{microtype}      
\usepackage{xcolor}         

\usepackage{wrapfig}
\usepackage{placeins} 
\usepackage{tikz}
\usetikzlibrary{arrows.meta,positioning,calc}
\usetikzlibrary{bayesnet}

\usepackage{graphicx}
\usepackage{subcaption}

\usepackage{hyperref}       

\usepackage{enumitem}
\usepackage{amsmath, amssymb, amsthm, mathtools}
\usepackage{bm}
\usepackage{bbm}

\usepackage{mdframed}
\newtheorem{proposition}{Proposition}

\newtheorem{assumption}{Assumption}

\DeclareRobustCommand{\newedit}[1]{\textcolor{orange}{#1}}

\definecolor{editpurple}{RGB}{120,40,200}

\usepackage[capitalize,noabbrev]{cleveref}
\definecolor{darkgreen}{rgb}{0.0, 0.7, 0.0}


\begin{document}

%
\runningtitle{Retrieval-Augmented Interpretable Learning}

%
\runningauthor{Mahbub et al.}

\twocolumn[
\aistatstitle{Retrieval-Augmented Interpretable Learning: \\ Towards Task-Specific Zero-Shot Models in Healthcare}
%
\aistatsauthor{
Sazan Mahbub$^{c}$ \quad
Caleb N. Ellington$^{c,g}$ \quad
Zhiyuan Li$^{w}$ \quad
Yixin Yang$^{w}$ \\
\textbf{Souvik Kundu}$^{i}$ \quad
\textbf{Ben Lengerich}$^{w}$ \quad
\textbf{Eric P. Xing}$^{c,m,g}$
}
\vspace{0.5em}
\aistatsaddress{
$^{c}$Carnegie Mellon University \quad
$^{w}$University of Wisconsin--Madison \\
$^{m}$Mohamed bin Zayed University of AI \quad
$^{g}$GenBio AI \quad
$^{i}$Intel
} ]

\noindent\textcolor{red}{\textit{NOTE: To avoid confusion, we note that a preliminary, non-archival workshop version of this work is available online under the name \textbf{RAG-IM}. RAIL is the renamed and completed version of the same work. This manuscript supersedes that earlier draft and should be consulted for the current formulation, results, and contributions.}}\\

\begin{abstract}
{We introduce Retrieval-Augmented Interpretable Learning (RAIL), a probabilistic meta-learning framework for zero-shot generation of task-specific interpretable models that synthesizes coefficient-space structure from natural-language task descriptions and a memory of previously learned task-specific predictors. RAIL retrieves related source tasks, transfers structure through coefficient space, and generates a new predictor in the original diagnostic-feature space, enabling zero-shot and few-shot clinical procedure prediction with feature-level explanations. Its probabilistic formulation provides uncertainty over retrieval, model coefficients, and predictions, supporting uncertainty-aware deployment: uncertain predictions or unstable explanations can be flagged for additional clinical review rather than treated as automatic decisions. This makes RAIL particularly suited for healthcare settings, where prediction tasks are highly long-tailed, new clinical targets arise frequently, and models must remain inspectable, uncertainty-aware, and compatible with human oversight. Across long-tailed clinical procedure prediction tasks, RAIL improves low-data model generation, benefits from clinically informed task representations, and yields retrieval, uncertainty, and coefficient-level diagnostics that make model behavior more transparent. These results suggest a path toward scalable clinical prediction systems that can adapt to new tasks while preserving interpretability and reliability.}
\end{abstract}

\section{Introduction}

\begingroup

Clinical prediction often requires separate models for many procedures, medications, or interventions, all conditioned on a shared set of diagnostic measurements. This setting is naturally long-tailed: common tasks have abundant labels, whereas clinically meaningful or newly emerging tasks may have only a few or no task-specific examples. Training each task independently therefore fails precisely where rapid adaptation is most needed. The challenge is not only to predict an outcome, but also to generate an appropriate and inspectable predictor for a new task under scarce supervision.

A growing body of work studies meta-models and task-conditioned predictors that generate or adapt models for new tasks or contexts~\citep{hospedales2021meta, lengerich2023contextualized, ellington2025learning, deuschel2024contextualizedpolicyrecoverymodeling}. Pretrained language models have also been explored as sources of semantic prior information for downstream decision-making~\citep{thirunavukarasu2023large}, including reinforcement learning~\citep{du2023guidingpretrainingreinforcementlearning, karimpanal2023lagrseqlanguageguidedreinforcementlearning, zhang2024adarefinerrefiningdecisionslanguage}, feature selection~{\citep{choi2022lmpriorspretrainedlanguagemodels}}, causal discovery~\citep{long2023causaldiscoverylanguagemodels, liu2024discoveryhiddenworldlarge}, and healthcare prediction pipelines~\citep{jin2024health}. However, many such systems produce latent representations, black-box predictors, or natural-language rationales rather than transparent task-specific models. Direct retrieval-transfer methods typically reuse or average existing predictors instead of synthesizing a task-specific parameter distribution. This leaves a gap between semantic task retrieval and intrinsically interpretable model generation. 

We ask whether an interpretable clinical predictor can be generated for a new task using only its natural-language description and a memory of previously learned interpretable predictors. We introduce \textbf{Retrieval-Augmented Interpretable Learning (RAIL)}, a probabilistic meta-learning framework for zero-shot and few-shot generation of task-specific predictors. Each memory entry pairs a semantic embedding of a task description with the parameters of its fitted interpretable predictor. For a target task, a learned retriever identifies relevant source tasks, and a probabilistic synthesizer produces a retrieval-conditioned distribution over the target predictor's parameters. The resulting predictor is deployed directly in the original diagnostic-feature space, so its feature-level weights remain inspectable without requiring a post-hoc explanation of an opaque model.

The central idea is that task descriptions and task-specific parameters encode complementary transferable knowledge. Semantic embeddings identify clinically related tasks, while source-model parameters capture how diagnostic features support their predictions. RAIL combines these signals through retrieval-conditioned coefficient synthesis, producing a target predictor in the zero-shot regime and allowing limited target-task labels to refine it in few-shot settings.

RAIL is not restricted to linear predictors. When an interpretable model class can be expressed through a shared basis, the same retrieval-conditioned synthesis mechanism can generate its parameters. We demonstrate this with RAIL$_{\mathrm{GAM}}$, {which synthesizes generalized additive models (GAMs)~\citep{hastie1986generalized} over a fixed cubic B-spline basis, producing} interpretable nonlinear shape functions while retaining zero-shot operation and coefficient uncertainty.

{On $7{,}487$ clinical procedure prediction tasks from MIMIC-IV, RAIL reaches pooled AUROC {$0.793$ both with only $2$--$4$ training examples and zero-shot}, with ECE at most $0.075$ across the low-data and zero-shot regimes, whereas task-specific logistic regression falls to AUROC $0.570$ and ECE $0.180$ with $2$--$4$ examples. On a separate patient- and era-disjoint benchmark, RAIL retains AUROC $0.755$--$0.758$.}

Our major contributions are as follows:
\begin{itemize}[leftmargin=1em, itemsep=0.01em, topsep=0.1em]
    \item We introduce \textbf{RAIL}, a probabilistic retrieval-conditioned meta-learning framework for zero-shot and few-shot generation of task-specific interpretable clinical predictors.
    \item We formulate model generation as learned retrieval-conditioned synthesis in a shared parameter space, pairing semantic task representations with fitted interpretable predictors and preserving direct feature-level inspection.
    \item We provide uncertainty over retrieval, synthesized parameters, and predictions, enabling failure detection, selective prediction, and evaluation of explanation stability.
    \item We show that the framework extends beyond linear models through RAIL$_{\mathrm{GAM}}$ and validate it on long-tailed clinical procedure prediction with supervised, retrieval-transfer, retrieval-free, and gradient-based references, calibration analyses, and patient- and era-disjoint generalization experiments.
\end{itemize}

\section{Related Work}
\label{sec:related_main}
{\textbf{Meta-learning and task-conditioned models.} Meta-learning uses experience across tasks to adapt to new tasks with limited data~\citep{hospedales2021meta,zhang2019metapred,tan2022metacare++}, and contextualized and task-conditioned models generate or adapt predictors for particular contexts or tasks~\citep{lengerich2019learning,lengerich2023contextualized,ellington2025learning,deuschel2024contextualizedpolicyrecoverymodeling}. RAIL shares this goal and generates the target predictor directly as an interpretable model in the original feature space, {from its task description and a memory of fitted predictors}.}

{\textbf{Retrieval augmentation.} Retrieval-augmented generation conditions a model on retrieved content~\citep{lewis2020retrieval,gao2023retrieval}, including health reports and medical knowledge for clinical prediction~\citep{jin2024health} and biological data for protein design~\citep{mahbub2025prism}. RAIL retrieves previously fitted predictors and reuses their coefficients as task-level structure.}

{\textbf{Language-model priors and interpretable models.} Pretrained language models supply prior knowledge when labels are scarce~\citep{thirunavukarasu2023large,choi2022lmpriorspretrainedlanguagemodels} and have been paired with interpretable models to keep the final decision transparent~\citep{bordt2024datasciencellmsinterpretable}. In RAIL, language-model embeddings relate tasks to one another, and the deployed predictor is an interpretable model with uncertainty over its coefficients. Extended related work is given in Appendix~\ref{sec:related_work}.}


\section{RAIL: \underline{R}etrieval-\underline{A}ugmented \underline{I}nterpretable \underline{L}earning}
\label{sec:method}
\begin{figure*}[t]
\centering
\includegraphics[width=0.64\textwidth]{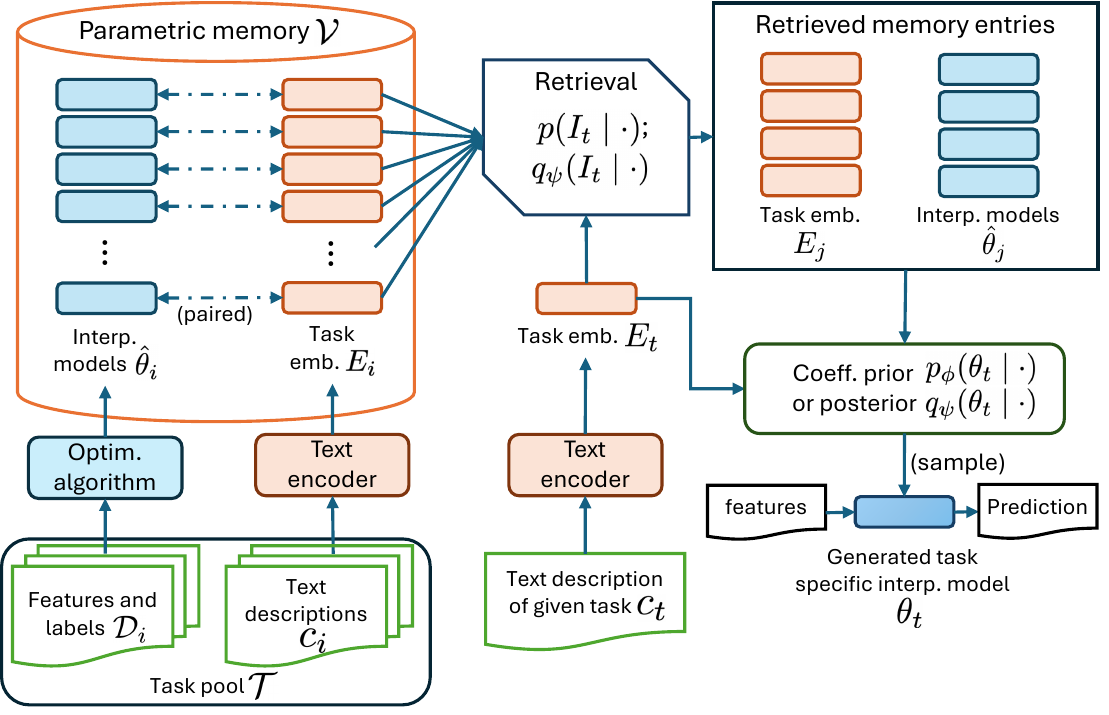} 
\caption{Overview of RAIL. Retrieved task embeddings and learned linear models from the parametric memory condition the generation of a task-specific interpretable predictor. 
}
\label{fig:rail_overall_architecture}
\vspace{-0.5em} 
\end{figure*}

\paragraph{Goal.}
Given a new task $t$ described only by natural language $c_t$, our goal is to synthesize an interpretable task-specific predictor
$f_t(x)=\sigma(\theta_t^\top {\Phi(x)})${, with a fixed feature map $\Phi$ (Sec.~\ref{sec:memory}),}
without training a new black-box model from scratch. RAIL assumes access to a memory of prior tasks, where each task is represented by both a language embedding and the coefficients of an interpretable linear model. 
The embeddings provide semantic alignment across tasks, while the coefficients provide transferable parameter-level structure in the original feature space. RAIL is primarily designed for zero-shot model synthesis: at inference time, no labeled examples from task $t$ are required. When limited target-task labels are available, the same formulation supports few-shot posterior adaptation by combining the retrieval-conditioned prior with a task-specific estimator.

\subsection{Parametric memory: semantic keys and interpretable values}
\label{sec:memory}


We assume training tasks $\{\mathcal T_i=(\mathcal D_i,c_i)\}_{i=1}^n$, where
$\mathcal D_i=\{(x_{ij},y_{ij})\}_{j=1}^{m_i}$ share a common feature space
$\mathcal X\subset\mathbb R^f$, and each task has a natural-language description $c_i$.
For each task $i$, we {fit an interpretable model that is linear in a fixed feature map $\Phi:\mathcal X\to\mathbb R^{d}$, $p(y=1\mid x)=\sigma\big(\theta^\top\Phi(x)\big)$, to obtain coefficients
$\hat\theta_i\in\mathbb R^{d}$ ($\Phi$ includes a constant $1$ for the intercept; explanations use the remaining coefficients).} We also compute a task embedding
$E_i=\mathrm{LM}(c_i)\in\mathbb R^e$ using a frozen language-model encoder. The resulting parametric memory is
\begin{align}
\mathcal V=\{(\hat\theta_i,E_i)\}_{i=1}^n ,
\end{align}
where embeddings serve as semantic retrieval keys and coefficients serve as interpretable parameter-space values.

\subsection{Latent-variable formulation}
    
RAIL introduces two latent quantities. {Cosine similarity first selects a candidate set $\mathcal C_t\subset\mathcal V$ of size $S$. The retrieval latent $I_t\in\{1,\ldots,S\}$ indexes an entry in this set; its categorical probabilities provide differentiable weights over the candidates in the implemented generator.} The task-parameter latent
{$\theta_t\in\mathbb R^{d}$} is the coefficient vector of the interpretable predictor for task $t$.

\subsection{Generative model}

The parametric memory $\mathcal V$ and task description $c_t$ are observed conditioning variables. RAIL models uncertainty over both {the retrieval index} and the task-specific coefficients:
\begin{align}
p_\phi(\mathcal D_t,\theta_t,I_t\mid c_t,\mathcal V)
&=
p(\mathcal D_t\mid \theta_t)\;
p_\phi(\theta_t\mid I_t,c_t) \nonumber\\
&\quad\times p(I_t\mid c_t,\mathcal V).
\label{eq:main-joint}
\end{align}
{The overall architecture is shown in Fig.~\ref{fig:rail_overall_architecture}, and the probabilistic graphical model in Fig.~\ref{fig:pgm-rail}.}
\begin{figure}[!htb]
\vspace{-0.1em}
\centering
\resizebox{0.36\textwidth}{!}{%
\begin{tikzpicture}[
    node distance=3.0cm and 4.6cm,
    rv/.style={
        circle,
        draw,
        thick,
        minimum size=17mm,
        inner sep=0pt,
        align=center,
        font=\fontsize{30}{30}\selectfont
    },
    obs/.style={
        rv,
        fill=gray!20
    },
    arrow/.style={
        -latex,
        thick,
        shorten >=2pt,
        shorten <=2pt
    }
]

\node[obs] (V) {$\mathcal{V}$};
\node[rv, below=2.7cm of V] (I) {$I_t$};

\node[rv, right=5.0cm of I] (theta) {$\theta_t$};
\node[obs, above=2.7cm of theta] (c) {$c_t$};

\node[obs, right=5.0cm of theta] (D) {$\mathcal{D}_t$};

\draw[arrow] (V) -- (I);

\draw[arrow] (c) -- (I);
\draw[arrow] (c) -- (theta);

\draw[arrow] (I) -- (theta);
\draw[arrow] (theta) -- (D);

\end{tikzpicture}%
}
\vspace{-0.2em}
\caption{
PGM for RAIL. Observed variables are shaded.
}
\label{fig:pgm-rail}
\end{figure}

\paragraph{Likelihood.}
For binary prediction, we use the logistic GLM likelihood
\begin{align}
p(y\mid x,\theta)=\mathrm{Bernoulli}\big({y;\,}\sigma(\theta^\top {\Phi(x)})\big),
\end{align}
where $(x,y)\in\mathcal D_t$. For regression, the likelihood can analogously be written as
$p(y\mid x,\theta)=\mathcal N(y;\theta^\top {\Phi(x)},{s^2})$.

\paragraph{Retrieval prior.}
The prior $p(I_t\mid c_t,\mathcal V)$ is induced by cosine similarity between the target task embedding $E_t=\mathrm{LM}(c_t)$ and memory entries. Let $s^{\mathrm{prior}}_{t,j}$ denote the cosine-similarity score for memory entry $j$. {We select the candidate set $\mathcal C_t$ of size $S$ using these scores and define normalized prior weights over its entries:}
\begin{align}
\pi^{\mathrm{prior}}_{t,j}
=
\frac{\exp(s^{\mathrm{prior}}_{t,j}/\tau_p)}
{\sum_{\ell=1}^{S}\exp(s^{\mathrm{prior}}_{t,\ell}/\tau_p)} .
\label{eq:retrieval-prior}
\end{align}
{Conditioned on this candidate set, the retrieval prior is $p(I_t\mid c_t,\mathcal V)=\mathrm{Cat}(\pi_t^{\mathrm{prior}})$.}
This cosine-based prior provides a semantic anchor for the learned retriever.

\paragraph{Retrieval-conditioned coefficient prior.}
{For each candidate memory entry $(\hat\theta_j,E_j)\in\mathcal C_t$, define}
\begin{align}
K_j=(\hat\theta_j\,\|\,E_j),
\qquad
V_j=\hat\theta_j .
\end{align}
Conditioned on $I_t$ and $c_t$, a multi-head cross-attention generator aggregates the {candidate} coefficient values and outputs a Gaussian prior over the target-task coefficients:
\begin{align}
p_\phi(\theta_t\mid I_t,c_t)
=
\mathcal N\!\big(
\theta_t;\mu_{\phi,t},\mathrm{diag}(\sigma^2_{\phi,t})
\big),
\label{eq:theta-prior}
\end{align}
where $(\mu_{\phi,t},\sigma^2_{\phi,t})$ are produced by neural heads applied to the attention output~\citep{vaswani2017attention}. We use a diagonal covariance and parameterize $\log\sigma^2_{\phi,t}$.

\subsection{Learned retrieval and coefficient inference}

We use the factorized variational family
\begin{align}
q_\psi(I_t,\theta_t\mid \mathcal D_t,c_t,\mathcal V)
=
q_\psi(I_t\mid c_t,\mathcal V)\;
q_\psi(\theta_t\mid \mathcal D_t,I_t,c_t).
\label{eq:q-factorization}
\end{align}
The retrieval factor $q_\psi(I_t\mid c_t,\mathcal V)$ is restricted to be label-free, so that the learned retriever can be used in zero-shot inference. It is trained jointly with the coefficient generator to improve over the cosine-similarity prior, while remaining semantically anchored to it through the retrieval KL term in Eq.~\ref{eq:neg-elbo}. The coefficient factor $q_\psi(\theta_t\mid \mathcal D_t,I_t,c_t)$ is the variational posterior used when target-task labels are available.

\paragraph{Amortized learned retriever.}
For each candidate memory entry, we compute a learned retrieval score from the target embedding and candidate key:
\begin{align}
s^{\mathrm{post}}_{t,j}=g_\psi(E_t,K_j).
\end{align}
Let $\mathcal K_t$ denote the top-$k$ candidates under these scores, with $k=|\mathcal K_t|<S$. To approximate near-discrete retrieval while preserving differentiability, we sharpen only positive top-$k$ scores:
\begin{align}
\tilde{s}^{\mathrm{post}}_{t,j}
=
\begin{cases}
s^{\mathrm{post}}_{t,j}/\tau_q, & j\in\mathcal K_t \text{ and } s^{\mathrm{post}}_{t,j}>0,\\
s^{\mathrm{post}}_{t,j}, & \text{otherwise},
\end{cases}
\label{eq:posterior-score-sharpening}
\end{align}
where $0<\tau_q\le 1$.
We then define the learned retrieval distribution
\begin{align}
q_\psi(I_t\mid c_t,\mathcal V)
&=
\mathrm{Cat}\!\big(\pi^{\mathrm{post}}_t\big), \nonumber\\
\pi^{\mathrm{post}}_{t,j}
&=
\frac{\exp(\tilde s^{\mathrm{post}}_{t,j})}
{\sum_{\ell=1}^{S}\exp(\tilde s^{\mathrm{post}}_{t,\ell})}.
\label{eq:retrieval-posterior}
\end{align}
{The learned probabilities ($\pi_t^{\mathrm{post}}$) multiplicatively modulate both the keys and values of the candidate memory entries before cross-attention. This emphasizes highly weighted candidates and attenuates the remaining entries, 
{approximating discrete retrieval over a smaller set while preserving differentiability.}}

\paragraph{Coefficient posterior.}
We use a Gaussian posterior with diagonal covariance:
\begin{align}
q_\psi(\theta_t\mid \mathcal D_t,I_t,c_t)
=
\mathcal N\!\big(
\theta_t;\mu_{q,t},\mathrm{diag}(\sigma^2_{q,t})
\big).
\label{eq:theta-posterior}
\end{align}
{For stability, we tie} the posterior and prior variances,
$\sigma^2_{q,t}\equiv\sigma^2_{\phi,t}$, and construct the posterior mean as
\begin{align}
\mu_{q,t}
=
\alpha\odot\mu_{\phi,t}
+
(1-\alpha)\odot\hat\theta^{\mathrm{self}}_t,
\label{eq:posterior-mean-gate}
\end{align}
where $\hat\theta^{\mathrm{self}}_t$ is an optional task-specific estimator, such as {the interpretable model of Sec.~\ref{sec:memory} fitted} on $\mathcal D_t$, and {$\alpha\in(0,1)^{d}$} is a learned per-coordinate gate. {The gate is global, shared across tasks, and optimized end-to-end. Thus, the posterior interpolates between the retrieval-conditioned prior and the task-specific estimator when target-task supervision is available.}

\paragraph{Untied posterior variance.}
{We also consider an untied variant, in which} the posterior precision {adds} target-task evidence per coordinate,
\begin{align}
\sigma_{q,t}^{-2}=\sigma_{\phi,t}^{-2}+n_t\,\Delta_t,\quad \Delta_t=\mathrm{softplus}\big(h_\psi(\hat\theta^{\mathrm{self}}_t)\big)\ge 0,
\end{align}
with $n_t=|\mathcal D_t|$, so the posterior variance contracts with more labels and equals the prior variance when $n_t=0$. {Appendix~\ref{app:untied_posterior} provides further methodological details and compares the tied and untied variants.}


\paragraph{Reparameterization and prediction.}
{When target-task labels are available, we sample}
\begin{align}
\theta_t
=
\mu_{q,t}
+
\sigma_{q,t}\odot\varepsilon,
\qquad
\varepsilon\sim\mathcal N(0,I),
\end{align}
and predict with the GLM likelihood, e.g., $\sigma(\theta_t^\top {\Phi(x)})$ for binary classification. {In zero-shot inference, we instead use the retrieval-conditioned prior parameters $\mu_{\phi,t}$ and $\sigma_{\phi,t}$. The prior mean defines a single deployable predictor, while coefficient samples support uncertainty-aware predictive averaging.}

\subsection{Variational objective with learned retrieval}

The task evidence marginalizes over both the retrieved memory index and the target-task coefficients:
\begin{align}
p_\phi(\mathcal D_t\mid c_t,\mathcal V)
=
\sum_{I_t}
\int
p_\phi(\mathcal D_t,\theta_t,I_t\mid c_t,\mathcal V)
\,d\theta_t .
\end{align}
{For the discrete-retrieval model, the restricted variational family in Eq.~\ref{eq:q-factorization} yields the following negative ELBO:}
\begin{mdframed}
\textbf{{Variational objective.}}
\begin{equation}
\begin{aligned}
&\mathcal J_t(\phi,\psi) = \\
&\quad -\mathbb E_{q_\psi(I_t\mid c_t,\mathcal V)}
\mathbb E_{q_\psi(\theta_t\mid \mathcal D_t,I_t,c_t)}
\big[
\log p(\mathcal D_t\mid \theta_t)
\big] \\
&\quad+
\mathbb E_{q_\psi(I_t\mid c_t,\mathcal V)}
\Big[
\mathrm{KL}\big(
q_\psi(\theta_t\mid \mathcal D_t,I_t,c_t) \\
&\qquad\qquad\qquad\qquad\big\|\,
p_\phi(\theta_t\mid I_t,c_t)
\big)
\Big] \\
&\quad+
\mathrm{KL}\big(
q_\psi(I_t\mid c_t,\mathcal V)
\,\big\|\,
p(I_t\mid c_t,\mathcal V)
\big).
\end{aligned}
\label{eq:neg-elbo}
\end{equation}
\end{mdframed}
{In implementation, we replace discrete-index sampling with soft retrieval weights in the coefficient generator. This yields a differentiable surrogate rather than an exact evaluation of the discrete-index expectation. The practical training objective and additional theoretical analyses are detailed in Appendix~\ref{app:additional-theory}.}
The likelihood term trains generated coefficients to explain target-task labels, the coefficient KL anchors the adapted posterior to the retrieval-conditioned prior, and the retrieval KL regularizes the learned retriever toward the semantic similarity prior. Thus, downstream prediction encourages retrieval of memory entries useful for coefficient synthesis, while the KL terms prevent unconstrained drift. 

\section{Results and Discussion}

{Our experiments assess:} (i) low-data and zero-shot predictive performance, (ii) the role of meaningful retrieval, (iii) whether task embeddings recover transferable parameter neighborhoods, and (iv) whether RAIL's uncertainty estimates support reliability and interpretability.

\subsection{Experimental Setup}\label{sec:setup}
\paragraph{Dataset.}
We construct clinical procedure prediction tasks from MIMIC-IV~\citep{johnson2023mimic}. Each task predicts whether a procedure is administered from $217$ diagnostic laboratory features, using the procedure {title} as the task description. The dataset contains $7{,}487$ procedure tasks: $4{,}412$ with at least two positive examples for supervised/few-shot evaluation and $3{,}075$ held-out tasks for zero-shot evaluation. {Dataset construction and evaluation protocols are detailed in Appendices~\ref{app:dataset_preprocessing} and~\ref{app:additional_experimental_results}.}

\paragraph{{Two benchmarks.}} {We evaluate on (i) the sample-scarcity benchmark and (ii) a separate patient- and era-disjoint benchmark, in which training and evaluation share no patients and cover different calendar eras, and the memory and RAIL are rebuilt from the training cohort (Sec.~\ref{sec:strict-benchmark}).}

\paragraph{Tasks and regimes.}
Each procedure defines a binary prediction task $t$ with diagnostic features $x\in\mathbb{R}^{f}$ (where $f=217$), a natural-language task description $c_t$, and task-specific labels. We evaluate across sample-scarcity regimes with $50+$, $10$--$49$, $5$--$9$, and $2$--$4$ {training examples per task (positive and negative)}, and a held-out zero-shot regime. In zero-shot evaluation, target tasks are completely excluded from the retrieval memory. 
RAIL receives only $c_t$, retrieves relevant source tasks distinct from the target task, and deploys the retrieval-conditioned coefficient prior for zero-shot inference.
{Each zero-shot task is evaluated on its single positive example and one negative example drawn from the positives of another randomly selected zero-shot task; negatives are fixed and shared by all methods (Appendix~\ref{app:dataset_preprocessing}).}

\paragraph{Baselines, ablations, and metrics.}
{We write RAIL$_{\mathrm{LR}}$ for the default instantiation that synthesizes logistic-regression predictors ($\Phi(x)=(1,x)$, so $d=f+1=218$) and RAIL$_{\mathrm{GAM}}$ for the instantiation that synthesizes GAMs (Sec.~\ref{sec:rail-gam}); unless stated otherwise, RAIL denotes RAIL$_{\mathrm{LR}}$.} We compare against supervised task-specific references, direct retrieval-transfer baselines, retrieval perturbations, and task-text encoder variants. Supervised references use target-task labels and are therefore not zero-shot methods. {They include logistic regression, GAMs, decision trees, MLPs, random forests~{\citep{breiman2001random}}, XGBoost~{\citep{chen2016xgboost}}, conditional hypernetwork~\citep{ha2017hypernetworks,Ba_2015_ICCV}, and the gradient-based meta-learning baselines, MAML$_{\mathrm{LR}}$ and MAML$_{\mathrm{MLP}}$, that apply second-order MAML~{\citep{finn2017model}} to a logistic-regression and an MLP base model, respectively. Both are meta-trained on the training splits of the memory tasks, the same information held in RAIL's memory, and adapt to a target task from its labeled examples; MAML$_{\mathrm{LR}}$ stays within RAIL's interpretable model class, while MAML$_{\mathrm{MLP}}$ serves as a non-interpretable reference.}
Retrieval-transfer baselines include Top-$1$ retrieved LR and {top-$100$} coefficient averaging. Retrieval ablations replace {learned} retrieval with random or semantically distant source tasks. For task-text representations, we compare DistilBERT-base-uncased{~\citep{sanh2019distilbert}} with the larger medical-domain MedEmbed-large-v0.1~{\citep{balachandran2024medembed}}. {We report pooled area under the ROC and precision--recall curves (AUROC, AUPRC), task-averaged accuracy, and calibration, measured by} expected calibration error (ECE)
{{\citep{naeini2015obtaining}}, maximum calibration error (MCE), and Brier score~{\citep{brier1950verification}}. 
A single temperature {(temperature scaling;~\citealp{guo2017calibration})} is fitted on the $50+$ evaluation split and transferred unchanged to the lower-resource and zero-shot regimes. {Tab.~\ref{tab:threshold-free-metrics} reports the low-data ($5$--$9$, $2$--$4$) and zero-shot regimes.} 


\subsection{Performance Across Sample-Scarcity Regimes}
{Procedure tasks are highly imbalanced in sample availability (Appendix Fig.~\ref{fig:long-tail}), which makes direct task-specific training unreliable in the low-data tail. Appendix Fig.~\ref{fig:long-tail-performance} 
compares RAIL with {task-specific LR} across {sample-scarcity} regimes. {Task-specific LR outperforms RAIL with abundant supervision ($50+$)} but degrades sharply as supervision decreases, from approximately $0.76$ to $0.42$ in F1 and from $0.76$ to $0.55$ in accuracy between the $50+$ and $2$--$4$ regimes. RAIL remains stable, with F1 between $0.70$ and $0.75$ and accuracy between $0.72$ and $0.75$, indicating that retrieval-conditioned model generation provides useful prior structure when direct supervision is scarce.} {Few-shot and zero-shot accuracies are close ($0.732$ vs.\ $0.734$) because both rely mainly on the retrieval-conditioned prior (Sec.~\ref{sec:method}).}


\subsection{Comparison with Baselines and Ablations}
\label{sec:baselines-ablations}
We compare RAIL with supervised references, a retrieval-free meta-learning baseline, retrieval perturbations, and direct retrieval-transfer baselines. 
{Tab.~\ref{tab:threshold-free-metrics} reports pooled discrimination and calibration against task-specific LR, and Tab.~\ref{tab:baseline-ablation} reports task-averaged accuracy for all baselines and ablations, together with whether each method exposes clinically useful diagnostics: an inspectable predictor, task-text conditioning, retrieval traceability, and coefficient uncertainty.}

\begin{table*}[!t]
\centering
\begingroup
\color{black}
\captionsetup{labelfont={color=black},textfont={color=black}}
\caption{{Discrimination and calibration in the low-data and zero-shot regimes. RAIL probabilities average $100$ coefficient samples; each method is temperature-scaled on the $50+$ split.} 
}

\label{tab:threshold-free-metrics}
\resizebox{0.75\textwidth}{!}{%
\begin{tabular}{lcccccccccc}
\toprule
Regime & \multicolumn{5}{c}{RAIL} & \multicolumn{5}{c}{Task-specific LR} \\
& AUROC & AUPRC & ECE & MCE & Brier & AUROC & AUPRC & ECE & MCE & Brier \\
\midrule
$5$--$9$   & {$0.813$} & {$0.765$} & $0.075$ & $0.133$ & $0.186$ & $0.642$ & $0.648$ & $0.100$ & $0.288$ & $0.249$ \\
$2$--$4$   & {$0.793$} & {$0.740$} & $0.068$ & $0.105$ & $0.193$ & $0.570$ & $0.567$ & $0.180$ & $0.369$ & $0.290$ \\
Zero-shot  & $0.793$ & $0.739$ & $0.067$ & $0.111$ & $0.194$ & -- & -- & -- & -- & -- \\
\bottomrule
\end{tabular}%
}
\endgroup
\vspace{-0.3em} 
\end{table*}




\begin{table*}[!t]
\centering
\caption{
Baseline and main ablation comparison. Diagnostic columns indicate whether a method provides an inspectable predictor, uses task-description embeddings, exposes retrieved source-task traces, or estimates coefficient uncertainty. {Entries with $\pm$ are means and standard deviations over five runs.
}
}
\label{tab:baseline-ablation}
\resizebox{0.9\textwidth}{!}{%
\begin{tabular}{lcccccc}
\toprule
Variant & Inspectable & Task text & Retrieval trace & Coeff. uncertainty & {Few-shot ($2$--$4$)} acc. & Zero-shot acc. \\
\midrule
\multicolumn{7}{l}{\emph{Supervised references}} \\

\quad {Task-specific LR}  
& \textcolor{darkgreen}{\checkmark} 
& \textcolor{red}{$\times$} 
& \textcolor{red}{$\times$} 
& \textcolor{red}{$\times$} 
& $0.550$ & N/A \\

{\quad Task-specific GAM}
& \textcolor{darkgreen}{\checkmark}
& \textcolor{red}{$\times$}
& \textcolor{red}{$\times$}
& \textcolor{red}{$\times$}
& {$0.576$} & {N/A} \\

\quad Task-specific Decision Tree 
& \textcolor{darkgreen}{\checkmark} 
& \textcolor{red}{$\times$} 
& \textcolor{red}{$\times$} 
& \textcolor{red}{$\times$} 
& $0.531$ & N/A \\

\quad Task-specific MLP 
& \textcolor{red}{$\times$} 
& \textcolor{red}{$\times$} 
& \textcolor{red}{$\times$} 
& \textcolor{red}{$\times$} 
& $0.533$ & N/A \\

\quad Task-specific XGBoost 
& \textcolor{red}{$\times$} 
& \textcolor{red}{$\times$} 
& \textcolor{red}{$\times$} 
& \textcolor{red}{$\times$} 
& $0.500$ & N/A \\

\quad Task-specific Random Forest 
& \textcolor{red}{$\times$} 
& \textcolor{red}{$\times$} 
& \textcolor{red}{$\times$} 
& \textcolor{red}{$\times$} 
& $0.573$ & N/A \\

\midrule
\multicolumn{7}{l}{\textcolor{black}{\emph{Gradient-based meta-learning {baselines}}}} \\

\textcolor{black}{\quad {MAML$_{\mathrm{MLP}}$}} 
& \textcolor{red}{$\times$}
& \textcolor{red}{$\times$}
& \textcolor{red}{$\times$}
& \textcolor{red}{$\times$}
& \textcolor{black}{$0.684$} & \textcolor{black}{N/A} \\

\textcolor{black}{\quad {MAML$_{\mathrm{LR}}$}} 
& \textcolor{darkgreen}{\checkmark}
& \textcolor{red}{$\times$}
& \textcolor{red}{$\times$}
& \textcolor{red}{$\times$}
& \textcolor{black}{$0.653$} & \textcolor{black}{N/A} \\

\midrule
\multicolumn{7}{l}{\emph{RAIL variants}} \\
\quad {RAIL$_{\mathrm{LR}}$} (w/ MedEmbed) 
& \textcolor{darkgreen}{\checkmark} 
& \textcolor{darkgreen}{\checkmark} 
& \textcolor{darkgreen}{\checkmark} 
& \textcolor{darkgreen}{\checkmark} 
& $0.732 \pm 0.009$ & $0.734 \pm 0.011$ \\

\quad {RAIL$_{\mathrm{LR}}$} (w/ DistilBERT) 
& \textcolor{darkgreen}{\checkmark} 
& \textcolor{darkgreen}{\checkmark} 
& \textcolor{darkgreen}{\checkmark} 
& \textcolor{darkgreen}{\checkmark} 
& $0.721 \pm 0.010$ & $0.718 \pm 0.013$ \\

\textcolor{black}{\quad RAIL$_{\mathrm{GAM}}$ (w/ MedEmbed){$^{\dagger}$}}
& \textcolor{darkgreen}{\checkmark}
& \textcolor{darkgreen}{\checkmark}
& \textcolor{darkgreen}{\checkmark}
& \textcolor{darkgreen}{\checkmark}
& \textcolor{black}{$0.728 \pm 0.008$} & \textcolor{black}{$0.724 \pm 0.014$} \\

\midrule
\multicolumn{7}{l}{\emph{Retrieval-free meta-learning baseline}} \\

\quad Text-conditioned hypernetwork
& \textcolor{darkgreen}{\checkmark} 
& \textcolor{darkgreen}{\checkmark} 
& \textcolor{red}{$\times$} 
& \textcolor{red}{$\times$} 
& $0.703 \pm 0.012$ & $0.701 \pm 0.014$ \\

\midrule
\multicolumn{7}{l}{\emph{Retrieval perturbation ablations}} \\

\quad RAIL + random retrieval 
& \textcolor{darkgreen}{\checkmark} 
& \textcolor{darkgreen}{\checkmark} 
& \textcolor{darkgreen}{\checkmark} 
& \textcolor{darkgreen}{\checkmark} 
& $0.657 \pm 0.019$ & $0.659 \pm 0.021$ \\

\quad RAIL + distant retrieval 
& \textcolor{darkgreen}{\checkmark} 
& \textcolor{darkgreen}{\checkmark} 
& \textcolor{darkgreen}{\checkmark} 
& \textcolor{darkgreen}{\checkmark} 
& $0.579 \pm 0.008$ & $0.587 \pm 0.010$ \\

\midrule
\multicolumn{7}{l}{\emph{Retrieval-transfer baselines}} \\

\quad Top-$1$ retrieved LR 
& \textcolor{darkgreen}{\checkmark} 
& \textcolor{darkgreen}{\checkmark} 
& \textcolor{darkgreen}{\checkmark} 
& \textcolor{red}{$\times$} 
& $0.587$ & $0.585$ \\

\quad {Top-$100$ coefficient average}
& \textcolor{darkgreen}{\checkmark} 
& \textcolor{darkgreen}{\checkmark} 
& \textcolor{darkgreen}{\checkmark} 
& \textcolor{red}{$\times$} 
& $0.656$ & $0.652$ \\

\bottomrule
\end{tabular} 
}

\vspace{2pt}
\begin{minipage}{0.9\textwidth}
\footnotesize
$^{\dagger}$RAIL$_{\mathrm{GAM}}$ {synthesizes GAMs over} a fixed cubic B-spline basis with $869$ coefficients while retaining the same retrieval, coefficient-synthesis, and uncertainty framework.
\end{minipage}
\vspace{-0.3em} 
\end{table*}

{{Across the low-data and zero-shot regimes, RAIL maintains AUROC between {$0.79$ and $0.81$}, ECE between $0.06$ and $0.08$, and Brier score between $0.18$ and $0.20$}, including zero-shot AUROC/AUPRC/ECE of $0.793/0.739/0.067${, whereas task-specific LR falls to AUROC $0.570$, ECE $0.180$, and Brier score $0.290$ with $2$--$4$ examples.
}

{With the same $2$--$4$ training examples per task, MAML$_{\mathrm{LR}}$ reaches accuracy $0.653$ and AUROC $0.702$, compared with $0.732$ and {$0.793$} for RAIL; 
the non-interpretable MAML$_{\mathrm{MLP}}$ reaches $0.684$ and $0.754$. Neither applies in the zero-shot regime.}

%

{Tab.~\ref{tab:baseline-ablation} shows that the RAIL variants provide the strongest combination of predictive performance and clinical inspectability.} MedEmbed improves over DistilBERT, suggesting that clinically informed task-text representations provide stronger priors for procedure prediction. The retrieval-free hypernetwork is competitive, showing that task descriptions carry useful signal; however, {RAIL} improves over it by grounding model generation in retrieved source-task coefficients and exposing a retrieval trace. Random and distant retrieval substantially degrade performance, showing that retrieval must be semantically meaningful. Direct retrieval-transfer baselines also underperform {RAIL}, indicating that learned coefficient synthesis is more effective than direct model reuse or {averaging of retrieved coefficients}. 


\begin{table}[!t]
\centering
\caption{{Zero-shot robustness: AUROC under feature masking (top) and accuracy with a random memory of $M$ tasks (bottom; $M<4{,}412$: mean over $10$ random subsets; $M=4{,}412$: full memory).}}
\label{tab:robustness-main}
{\footnotesize\setlength{\tabcolsep}{3pt}
{\begin{tabular}{lccccc}
\toprule
Masked features & $0\%$ & $20\%$ & $50\%$ & $80\%$ & $100\%$ \\
AUROC & $0.793$ & $0.738$ & $0.666$ & $0.603$ & $0.500$ \\
\midrule
Memory $M$ & $100$ & $500$ & $1{,}000$ & $2{,}000$ & $4{,}412$ \\
Accuracy & $0.704$ & $0.721$ & $0.728$ & $0.731$ & $0.734$ \\
\bottomrule
\end{tabular}}}
\vspace{-0.3em} 
\end{table}
\paragraph{Robustness.} Zero-shot AUROC degrades smoothly as increasing fractions of the shared feature space are masked {(Tab.~\ref{tab:robustness-main}; Appendix Tab.~\ref{tab:feature-masking-results}).} 
 Memory coverage also matters, yet $1{,}000$ source tasks already recover most of the full-memory performance {(Tab.~\ref{tab:robustness-main}; Appendix Tab.~\ref{tab:memory-size-ablation})}.

\subsection{Patient- and Era-Disjoint Benchmark}
\label{sec:strict-benchmark}
%
{On the second benchmark, the memory is refit on the training cohort and RAIL is retrained ($4{,}395$ memory tasks, $3{,}092$ zero-shot targets).}

\begin{table}[!t]
\centering
\caption{{Patient- and era-disjoint benchmark (training 2008--2013, evaluation 2014--2019).}}
\label{tab:strict-split-generalization}
{\small\setlength{\tabcolsep}{4pt}%
\begin{tabular}{lccc}
\toprule
Regime & \begin{tabular}[b]{@{}c@{}}RAIL\\AUROC\end{tabular} & \begin{tabular}[b]{@{}c@{}}Task-specific\\LR AUROC\end{tabular} & \begin{tabular}[b]{@{}c@{}}RAIL\\ECE\end{tabular} \\
\midrule
Few-shot ($2$--$4$) & $0.758$ & $0.580$ & $0.054$ \\
Zero-shot          & $0.755$ & --       & $0.060$ \\
\bottomrule
\end{tabular}}
\vspace{-0.3em} 
\end{table}


{{On this benchmark (Tab.~\ref{tab:strict-split-generalization}), RAIL retains AUROC $0.758$ in few-shot and $0.755$ in zero-shot evaluation, with ECE of $0.054$ and $0.060$, while task-specific LR falls to AUROC $0.580$ with $2$--$4$ examples, and unable to produce any prediction in zero-shot setting.}}

\subsection{Retrieval and Embedding Diagnostics}
{RAIL assumes that semantically related task descriptions retrieve source tasks with transferable coefficient structure. Nearest neighbors in task-embedding space are substantially closer to the query in coefficient space, and more internally coherent, than random neighbors (Appendix Fig.~\ref{fig:embedding-model-alignment}), and embedding retrieval achieves higher oracle overlap than random retrieval while matching or exceeding TF-IDF (Appendix Fig.~\ref{fig:oracle-retrieval}). 
Consistently, Gaussian noise on the task embeddings degrades few-shot and zero-shot performance, compressed (PCA) embeddings preserve useful retrieval structure, and random or shuffled embeddings collapse retrieval quality (Appendix Fig.~\ref{fig:app-embedding-ablation}). RAIL therefore depends on semantic organization in the task-embedding space rather than only on lexical overlap.} {Richer task descriptions help modestly: expanding each procedure title into a longer description {with an LLM (Appendix~\ref{app:description-expansion})} raises zero-shot AUROC from $0.793$ to $0.802$ (Appendix Tab.~\ref{tab:description-expansion-results}).}
\textbf{Near-neighbor removal.} {To test whether the gains come from copying near-duplicate procedures, we remove memory tasks at inference time on the {patient- and era-disjoint} benchmark (Tab.~\ref{tab:near-neighbor-main}; full controls in Appendix Tab.~\ref{tab:near-neighbor-overlap-controls}). Removing the $100$ nearest tasks
lowers AUROC by $0.0112$, whereas removing the $2{,}000$ nearest reduces it far more than {removing the same number of random tasks}, so RAIL relies on broad semantic coverage rather than a single nearest predictor. No zero-shot target description exactly matches a memory task ($0/3{,}092$), and removing tasks that share a three-character ICD-10-PCS prefix or the $100$ nearest TF-IDF descriptions changes AUROC by $0.0033$ and $0.0089$, respectively.}

\begin{table}[!t]
\centering
\caption{{Zero-shot near-neighbor controls on the patient- and era-disjoint benchmark (subset of Appendix Tab.~\ref{tab:near-neighbor-overlap-controls}).}}
\label{tab:near-neighbor-main}
{\footnotesize\setlength{\tabcolsep}{3pt}
\resizebox{\ifdim\width>\columnwidth\columnwidth\else\width\fi}{!}{
{\begin{tabular}{lcc}
\toprule
Memory operation & AUROC & AUPRC \\
\midrule
No removal & $0.7552$ & $0.7267$ \\
Remove nearest $100$ (MedEmbed) & $0.7440$ & $0.7187$ \\
{Random removal}, $100$ & $0.7550$ & $0.7269$ \\
Remove nearest $2{,}000$ (MedEmbed) & $0.7038$ & $0.6811$ \\
{Random removal}, $2{,}000$ & $0.7507$ & $0.7231$ \\
Remove nearest $100$ (TF-IDF) & $0.7463$ & $0.7198$ \\
{Remove shared $3$-char.\ PCS prefix} & $0.7519$ & $0.7229$ \\
\bottomrule
\end{tabular}}}}
\vspace{-0.1em} 
\end{table}

\vspace{-0.3em}
\subsection{Uncertainty Diagnostics and Selective Prediction}
\vspace{-0.2em}
RAIL provides retrieval, coefficient, and predictive uncertainty. We focus on predictive uncertainty for reliability analysis: {we sample $100$ coefficient vectors from the posterior (the prior in zero-shot), compute the induced predictive probabilities, and use the binary entropy of their mean $\bar p$ as the uncertainty score.} Prediction confidence is $\max(\bar p,1-\bar p)$.

Prediction confidence aligns with empirical correctness ({Appendix} Fig.~\ref{fig:uncertainty-basic}): accuracy increases monotonically 
from low- to high-confidence bins, while error rate increases from low- to high-uncertainty bins. 
We further define a sample-level failure as $\mathbf{1}[\hat y\ne y]$ and evaluate uncertainty scores using AUROC and AUPRC. Predictive uncertainty is the strongest failure-detection signal, achieving AUROC $0.683$ and AUPRC $0.419$, outperforming retrieval entropy and parameter uncertainty; ROC and precision--recall curves are provided in Appendix Fig.~\ref{fig:app-failure-detection}. Thus, output-level uncertainty is most aligned with prediction-level failure, while retrieval and coefficient uncertainty {support source attribution and explanation stability (Sec.~\ref{sec:rail-gam})}.

\begin{figure}[t]
\centering
\includegraphics[width=0.75\columnwidth]{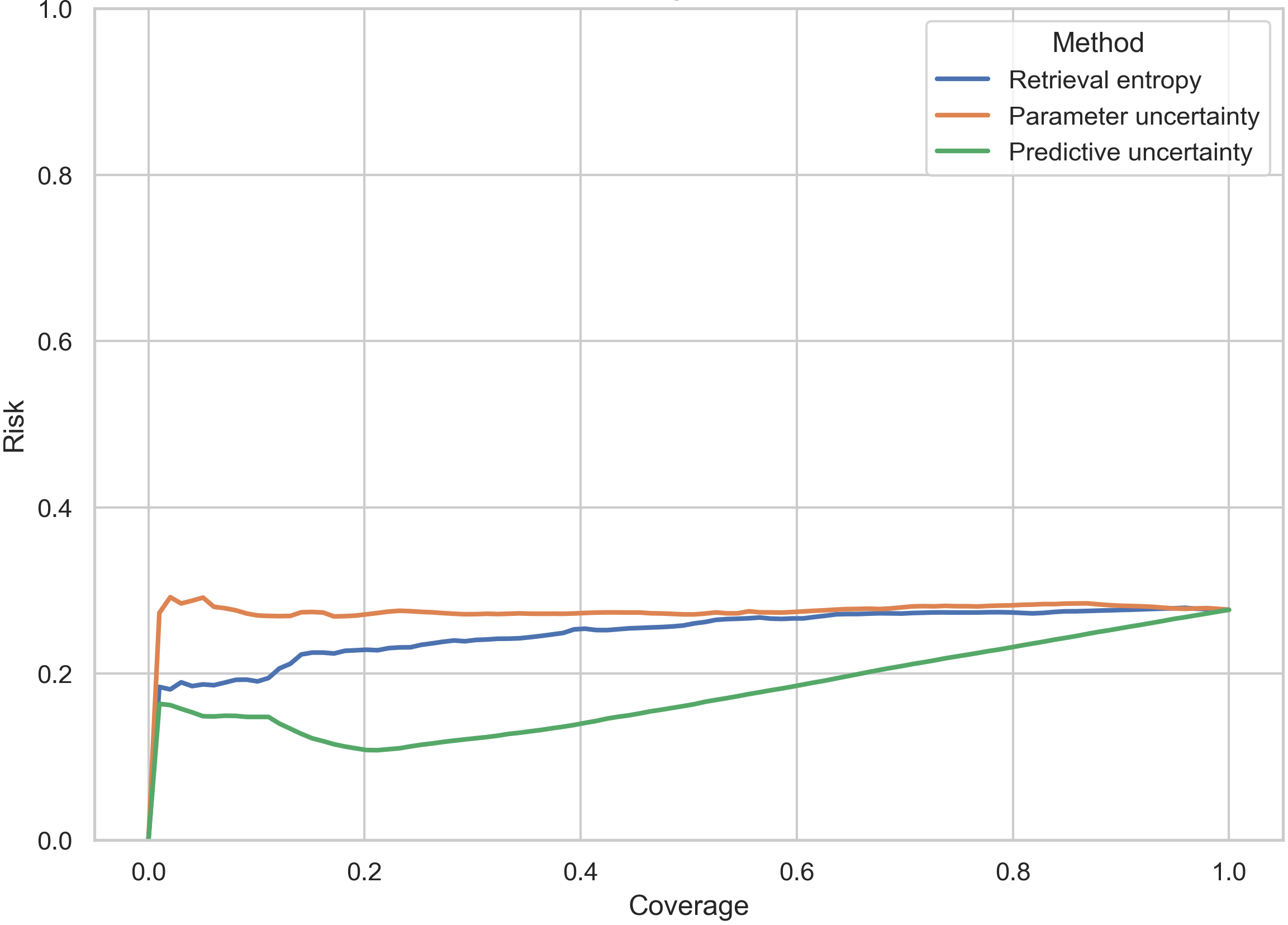}
\caption{{Selective prediction (risk vs.\ coverage) for three uncertainty signals; ranking by predictive uncertainty lowers risk as coverage decreases.}}
\label{fig:selective-prediction}
\vspace{-0.5em} 
\end{figure}
Finally, we evaluate uncertainty-based abstention by ranking examples by increasing uncertainty and retaining the lowest-uncertainty subset. Fig.~\ref{fig:selective-prediction} shows that risk decreases as coverage decreases, suggesting that uncertain predictions can be flagged for expert review rather than used automatically.

{A separately trained model with an untied, evidence-dependent posterior variance improves AUROC ($0.796/0.799$ few-/zero-shot) and Brier score, and its posterior standard deviation shrinks to $0.575$ times the prior standard deviation
with $5{,}000$ target-task examples 
(see details in Appendix~\ref{app:untied_posterior}).}

\subsection{Interpretability and Extensions}


For each task and feature, we record posterior coefficient mean, magnitude, variance, and standard deviation ({Appendix} Fig.~\ref{fig:feature-ci}). Stable features have large posterior importance with narrow intervals, while uncertain features have wider intervals and should be interpreted cautiously. Additional stable and unstable coefficient examples are provided in Appendix Fig.~\ref{fig:app-stable-unstable-coefs}.

\textbf{Generality beyond linear models.}\label{sec:rail-gam} {Because RAIL synthesizes coefficients in a shared space, it extends to any interpretable model class with a fixed shared basis. RAIL$_{\mathrm{GAM}}$ {synthesizes GAMs over} a cubic B-spline basis, which expands the coefficient space from {$218$} to $869$ dimensions and yields directly inspectable nonlinear shape functions (Tab.~\ref{tab:interpretable-model-classes}; details in Appendix Tab.~\ref{tab:gam-implementation}).}

\begin{table}[!t] 
\centering
\caption{{Pooled AUROC and AUPRC across interpretable model classes (FS: $2$--$4$ training examples; ZS: zero-shot).}}
\label{tab:interpretable-model-classes}
{\small\setlength{\tabcolsep}{4pt}%
\begin{tabular}{lcccc}
\toprule
Metric & {\begin{tabular}[b]{@{}c@{}}Per-task\\LR\end{tabular}} & \begin{tabular}[b]{@{}c@{}}Per-task\\GAM\end{tabular} & {RAIL$_{\mathrm{LR}}$} & {RAIL$_{\mathrm{GAM}}$} \\
\midrule
FS AUROC  & {$0.570$} & $0.600$ & {$0.793$} & $0.802$ \\
FS AUPRC  & {$0.567$} & $0.576$ & {$0.740$} & $0.760$ \\
ZS AUROC  & \newedit{--}      & --      & $0.793$ & $0.801$ \\
ZS AUPRC  & \newedit{--}      & --      & $0.739$ & $0.766$ \\
\bottomrule
\end{tabular}}
\vspace{-0.3em} 
\end{table}

{{RAIL$_{\mathrm{GAM}}$ improves over RAIL$_{\mathrm{LR}}$ in pooled AUROC/AUPRC by $+0.008/+0.027$ in zero-shot evaluation, while its task-averaged accuracy stays within $0.010$ of RAIL$_{\mathrm{LR}}$ (Tab.~\ref{tab:baseline-ablation}).}
} 

\vspace{-1em}
\subsection{Scope, Limitations, and Future Work}
\label{sec:limitations}
{RAIL relies on a shared diagnostic-feature space and on memory coverage of related source tasks, and its coefficient explanations are associational rather than causal. Extending RAIL to new feature identities and validating it across institutions and prospectively are natural next steps.}

\vspace{-1em}
\section{Conclusion}
\vspace{-1em}
We introduced \textbf{RAIL}, a probabilistic meta-learning framework that retrieves relevant source tasks and synthesizes a distribution over the parameters of a task-specific interpretable clinical predictor. By operating directly on the original diagnostic features, RAIL preserves intrinsic interpretability while quantifying uncertainty over retrieval, parameters, and predictions. RAIL addresses settings with scarce or absent supervision, achieving a zero-shot AUROC of $0.793$ without target-task training labels. Evaluations under patient- and era-disjoint splits show good calibration, and extensions to nonlinear GAMs demonstrate the framework's flexibility. Additional diagnostics support uncertainty-aware selective prediction and show robustness to missing features, smaller memories, and the removal of near-duplicate source tasks. More broadly, RAIL offers a path toward clinical prediction systems that translate descriptions of new or rare tasks into inspectable models, with explicit uncertainty to help identify cases requiring clinician review.
\subsection*{{AI Use Statement}}
{In this work, we used generative AI tools to generate expanded task descriptions for one sensitivity analysis (Appendix~\ref{app:description-expansion}): OpenAI's \texttt{gpt-4.1-mini} rewrote each ICD-10-PCS procedure title into a longer description. Each request contained only the public procedure title; no MIMIC-IV patient-level data were sent to any external service. All main results use the original procedure titles. 
Additionally, we used generative AI tools to edit the paper for readability.
}



\bibliographystyle{plainnat}
\bibliography{iclr2027_conference}

\section*{Checklist}

\begin{enumerate}

  \item For all models and algorithms presented, check if you include:
  \begin{enumerate}
    \item A clear description of the mathematical setting, assumptions, algorithm, and/or model. 
    {Yes.}
    \item An analysis of the properties and complexity (time, space, sample size) of any algorithm. 
    {Yes.}
    \item (Optional) Anonymized source code, with specification of all dependencies, including external libraries. No, code will be released upon acceptance.
  \end{enumerate}

  \item For any theoretical claim, check if you include:
  \begin{enumerate}
    \item Statements of the full set of assumptions of all theoretical results. {Yes.}
    \item Complete proofs of all theoretical results. {Yes.}
    \item Clear explanations of any assumptions. {Yes.}
  \end{enumerate}

  \item For all figures and tables that present empirical results, check if you include:
  \begin{enumerate}
    \item The code, data, and instructions needed to reproduce the main experimental results (either in the supplemental material or as a URL). {No, code will be released upon acceptance; MIMIC-IV is available through PhysioNet credentialed access; dataset construction is described in Appendix~\ref{app:dataset_preprocessing}.}
    \item All the training details (e.g., data splits, hyperparameters, how they were chosen). Yes.
    \item A clear definition of the specific measure or statistics and error bars (e.g., with respect to the random seed after running experiments multiple times). Yes.
    \item A description of the computing infrastructure used. (e.g., type of GPUs, internal cluster, or cloud provider). 
    Yes.
  \end{enumerate}

  \item If you are using existing assets (e.g., code, data, models) or curating/releasing new assets, check if you include:
  \begin{enumerate}
    \item Citations of the creator If your work uses existing assets. 
    Yes.
    \item The license information of the assets, if applicable. 
    Yes.
    \item New assets either in the supplemental material or as a URL, if applicable. Not Applicable.
    \item Information about consent from data providers/curators. 
    Yes.
    \item Discussion of sensible content if applicable, e.g., personally identifiable information or offensive content. {Yes, The data are de-identified, and no patient-level data were sent to external services (Appendix~\ref{app:description-expansion}).}
  \end{enumerate}

  \item If you used crowdsourcing or conducted research with human subjects, check if you include:
  \begin{enumerate}
    \item The full text of instructions given to participants and screenshots. Not Applicable.
    \item Descriptions of potential participant risks, with links to Institutional Review Board (IRB) approvals if applicable. Not Applicable.
    \item The estimated hourly wage paid to participants and the total amount spent on participant compensation. Not Applicable.
  \end{enumerate}

\end{enumerate}

\clearpage
\appendix
\thispagestyle{empty}

\onecolumn
\aistatstitle{Retrieval-Augmented Interpretable Learning:\\ Towards Task-Specific Zero-Shot Models in Healthcare.\\ Supplementary Materials}

\section{\texorpdfstring{{Extended Related Work}}{Extended Related Work}}
\label{sec:related_work}

Clinical predictive modeling often faces a data-scarcity problem: many clinically relevant tasks have limited supervision, yet the resulting models must remain accurate, interpretable, and reliable enough to support expert review.\footnote{A preliminary, non-archival workshop version of this work (RAIL) exists under a different name (RAG-IM); it does not constitute prior publication. It is an earlier draft of this same work, not a separate method. All of its content is superseded by this paper.}
Several research directions address different parts of this challenge. Contextualized machine learning studies prediction settings where the input--output relationship varies across contexts, individuals, or tasks, which is common in clinical and decision-making domains~\citep{lengerich2023contextualized, deuschel2024contextualizedpolicyrecoverymodeling, lengerich2019learning,ellington2025learning}.
Meta-learning similarly seeks to use experience across tasks to support rapid adaptation to new tasks with limited data~\citep{zhang2019metapred,tan2022metacare++,hospedales2021meta}{, with MAML~\citep{finn2017model} a widely used gradient-based instance}. Recent meta-models and task-conditioned predictors extend this idea by dynamically generating or adapting predictors for particular contexts or tasks~\citep{hospedales2021meta, lengerich2023contextualized, ellington2025learning, deuschel2024contextualizedpolicyrecoverymodeling}. These methods provide a foundation for scalable low-data prediction, but many rely on latent representations or flexible neural predictors rather than directly inspectable models in the original feature space.

Another approach is to use retrieval to provide task-relevant external information. Retrieval-augmented generation conditions a model on retrieved content, allowing it to use information beyond what is stored in its parameters~\citep{lewis2020retrieval, gao2023retrieval}. 
{In the biomedical domain}, retrieval-augmented pipelines have been used to incorporate health reports and medical knowledge into large models for improved feature extraction and prediction~\citep{jin2024health}{, and retrieval over biological data has been used for scientific problems such as protein design~\citep{mahbub2025prism}}. Retrieval is therefore a natural mechanism for grounding predictions in relevant prior information. However, standard RAG systems typically retrieve documents or knowledge snippets for language generation or latent prediction, rather than using retrieved prior predictors as reusable task-level structure.

A complementary direction uses pretrained language models {for adaptation in data-scarce settings}~\citep{thirunavukarasu2023large,mahbub2022bioadapt}. 
Language-model priors have been used in reinforcement learning to guide exploration and decision refinement~\citep{du2023guidingpretrainingreinforcementlearning, karimpanal2023lagrseqlanguageguidedreinforcementlearning, zhang2024adarefinerrefiningdecisionslanguage}, in feature selection to identify useful variables~{\citep{choi2022lmpriorspretrainedlanguagemodels}}, and in causal discovery or reasoning to propose or evaluate plausible causal structure~\citep{long2023causaldiscoverylanguagemodels, kiciman2024causal, liu2024discoveryhiddenworldlarge}. Related work has also studied language-model priors across multiple decision-making settings~\citep{choi2022lmpriorspretrainedlanguagemodels,li2022pre,yan2025efficient}{, and transformer sequence models have been used to cast reinforcement learning as sequence modeling~\citep{chen2021decisiontransformerreinforcementlearning}}. These studies show that pretrained representations can encode useful task-level information, but this information is often used to guide black-box decision processes, construct latent features, or generate natural-language rationales rather than directly producing task-specific interpretable predictors.

Finally, work combining language models with interpretable models aims to leverage language-derived knowledge while preserving transparency in the final decision process~\citep{bordt2024datasciencellmsinterpretable}. This direction is especially relevant for healthcare, where users may need to inspect feature-level evidence, assess reliability, and decide when additional review is needed{~\citep{tonekaboni2019clinicians}}. Overall, prior work offers important tools for low-data prediction---task adaptation, retrieval, language-derived priors, and interpretability---but leaves open the challenge of combining these ingredients into a framework that produces task-specific predictors that are simultaneously data-efficient, directly inspectable, and uncertainty-aware.

\begin{figure}[!htb]
\centering
\includegraphics[width=0.76\textwidth]{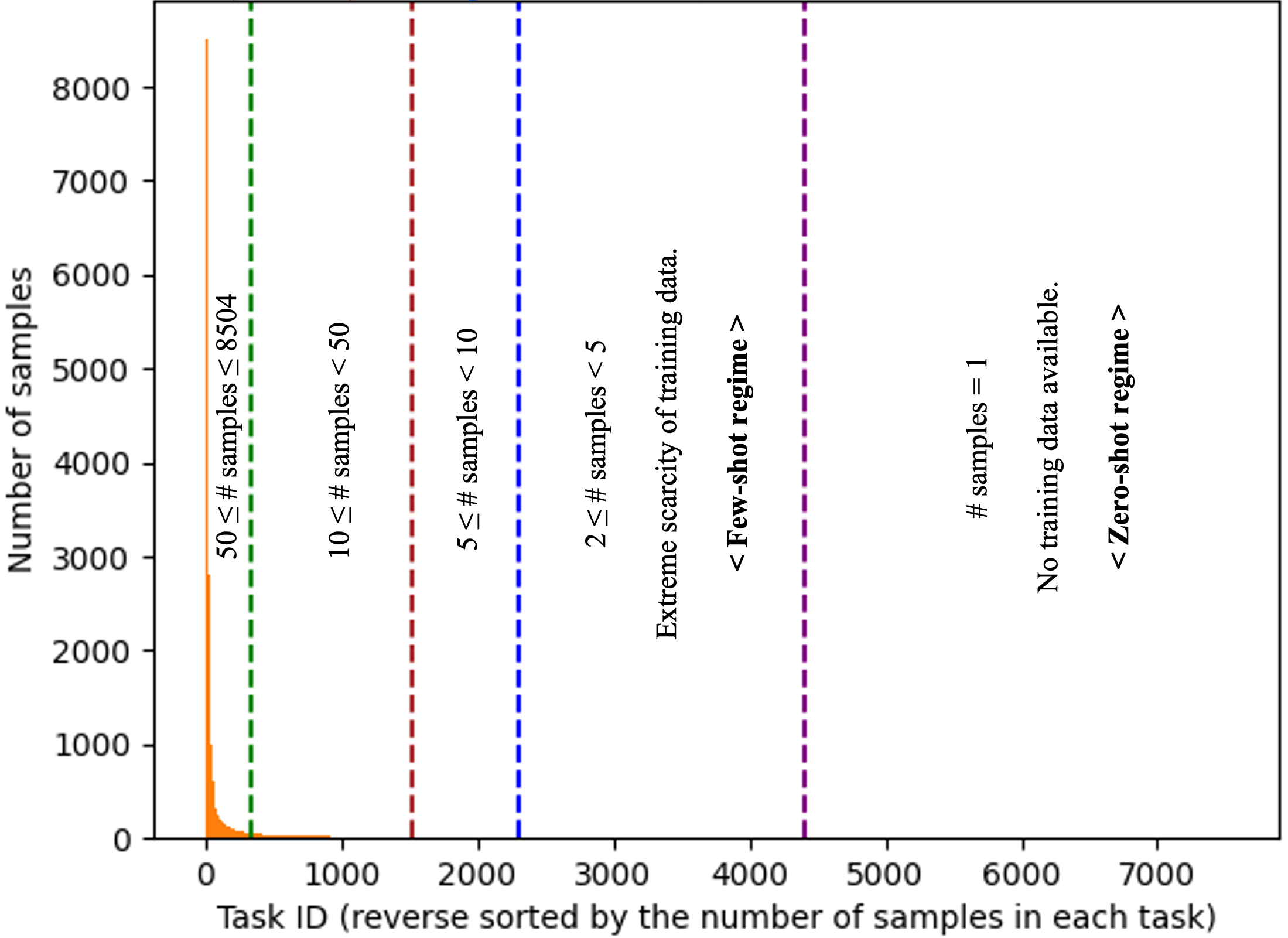}
\caption{
Long-tailed task distribution. Procedure tasks are highly imbalanced in sample availability. 
}
\label{fig:long-tail}
\vspace{-0.6em} 
\end{figure}



\section{Dataset and Preprocessing}
\label{app:dataset_preprocessing}

We use the MIMIC-IV dataset~\citep{johnson2023mimic} to construct clinical procedure prediction tasks. The preprocessed data consist of two main tables: laboratory events and procedure records. Both tables are grouped by hospital admission and patient identifiers, $(\texttt{hadm\_id}, \texttt{subject\_id})$. Laboratory events provide diagnostic measurements, which form the input features, while procedure records define task-specific binary labels indicating whether a procedure was administered during a hospital stay. We retain only admissions with corresponding laboratory measurements so that every labeled example has an associated diagnostic feature vector.

Each input vector contains $217$ diagnostic features derived from laboratory measurements. Since laboratory items have different clinical reference ranges, we normalize each raw measurement using its item-specific lower and upper reference values:
\begin{equation}
    \label{eq:normalization}
    \mathrm{normalized\_feature}
    =
    \frac{\mathrm{raw\_value}-\mathrm{lower\_range}}
    {\mathrm{upper\_range}-\mathrm{lower\_range}} .
\end{equation}
Values outside the reference interval are not clipped; this preserves clinically meaningful deviations while placing heterogeneous diagnostics on a comparable scale.

Each clinical procedure defines a separate binary prediction task. The dataset contains $7{,}487$ procedure tasks in total. Among these, $4{,}412$ procedures have at least two positive examples and are used for high-resource, low-resource, and few-shot evaluation. For each of these tasks, we construct balanced task-specific datasets with equal numbers of positive and negative examples and use a $50/50$ train--test split within the task. The remaining $3{,}075$ procedure tasks are used for zero-shot evaluation, where no target-task training examples or task-specific models are available during inference. {Each zero-shot task has a single positive example. Its test set pairs this example with one negative example, drawn from the positive examples of another randomly selected zero-shot task, so each zero-shot task is evaluated on one positive and one negative example ($6{,}150$ test examples in total). We verified that no sampled negative's admission includes the target procedure. {Negatives were sampled once and kept fixed, so all methods and runs are evaluated on the same test pairs.}} {The same negative sampling is used for the supervised and few-shot regimes: the positive examples of a task are split evenly between its training and test sets, and each set is completed with an equal number of negative examples drawn from the positive examples of other tasks. Regime sizes count all training examples of a task, positive and negative; for example, a task with two positive examples has one positive and one negative training example (the $2$--$4$ regime) and a test set of the same size.} {During training, a task's own memory entry is excluded when it is the target; at few-shot inference, the memory contains all source tasks, including the target's own entry, which is fitted on its training split only. MIMIC-IV is a de-identified database accessed under the PhysioNet Credentialed Health Data License and data use agreement 
Tab.~\ref{tab:task-regimes} summarizes the resulting task regimes.

We use {procedure titles} as task descriptions. Example descriptions include ``Respiratory Ventilation, 24--96 Consecutive Hours,'' ``Inspection of Upper Intestinal Tract, Via Natural or Artificial Opening Endoscopic,'' ``Fluoroscopy of Multiple Coronary Arteries using Other Contrast,'' and ``Introduction of Other Antineoplastic into Central Vein, Percutaneous Approach.'' These descriptions provide the task-text signal used by RAIL for retrieval-augmented model generation.

\begin{table}[h]
\centering
\caption{
Summary of task regimes used in RAIL evaluation. Each clinical procedure defines one binary prediction task. The zero-shot regime contains held-out target procedures whose labels and task-specific models are excluded from the retrieval memory during inference. {Examples per task count the training examples of a task, positive and negative.}
}
\label{tab:task-regimes}
\begin{tabular}{lll}
\toprule
Regime & Task ID range & Examples per task \\
\midrule
High-resource & $0$--$341$ & $50+$ \\
Moderate-resource & $342$--$1521$ & $10$--$49$ \\
Low-resource & $1522$--$2306$ & $5$--$9$ \\
Few-shot & $2307$--$4411$ & $2$--$4$ \\
Zero-shot & $4412$--$7486$ & No target-task labels used \\
\bottomrule
\end{tabular}
\vspace{-0.4em} 
\end{table}

\section{Hardware Specifications}
\label{apd:hardware_spec}

All experiments were run on a single workstation equipped with 12 CPU cores of AMD EPYC 9354 processor at 3.25 GHz, 100 GB of RAM, and one NVIDIA RTX A6000 GPU with 48 GB of VRAM.

\section{Additional Implementation Details and Hyperparameters}
\label{app:hyperparameters}
{Tab.~\ref{tab:hyperparameters} lists the settings used for the main RAIL results. The training objective is the weighted surrogate in Eq.~\ref{eq:loss-practical}. 

\begin{table}[!ht]
\centering
\caption{Implementation settings for RAIL. }
\label{tab:hyperparameters}
\small
{%
\begin{tabular}{@{}p{0.17\textwidth}p{0.42\textwidth}p{0.33\textwidth}@{}}
\toprule
Component & Setting & Value \\
\midrule
Data & Diagnostic features $f$ & {$217$ ($+$ intercept)} \\
 & Memory (source) tasks $n$ / zero-shot tasks & $4{,}412$ / $3{,}075$ \\
 & {Patient- and era-disjoint benchmark} & {$4{,}395$ / $3{,}092$} \\
Task encoder & Text encoder (main / ablation) & {MedEmbed-large-v0.1} / DistilBERT-base-uncased \\
 & Encoder frozen or fine-tuned & {Frozen (Sec.~\ref{sec:method})} \\
Retrieval & Candidate set size $S$ & $100$ \\
Inference and & {Monte Carlo coefficient samples} & $100$ \\
calibration & Temperature (fitted on the $50+$ split) & $\approx 5.15$ (tied) / $\approx 3.55$ (untied)\\
 & ECE bins & $15$ equal-width \\
RAIL$_{\mathrm{GAM}}$ & Basis & cubic B-spline; $4$ coefficients per feature; quantile knots \\
 & Synthesized coefficients & $217\times 4+1=869$ \\
\bottomrule
\end{tabular}}
\vspace{-0.4em} 
\end{table}

\section{Additional Retrieval Robustness, Uncertainty, and Interpretability Diagnostics}
\label{app:additional-robustness-uncertainty}

{This section collects additional diagnostic figures: {oracle} retrieval consistency (Fig.~\ref{fig:oracle-retrieval}), feature-level coefficient uncertainty (Fig.~\ref{fig:feature-ci}), embedding--model alignment (Fig.~\ref{fig:embedding-model-alignment}), and performance across sample-scarcity regimes (Fig.~\ref{fig:long-tail-performance}). Representation ablations, failure-detection curves, and stable versus unstable coefficient examples are in Appendix~\ref{app:additional_experimental_results}.}


\begin{figure}[!htb]
    \vspace{-0.0em}
    \centering
    \includegraphics[width=0.58\textwidth]{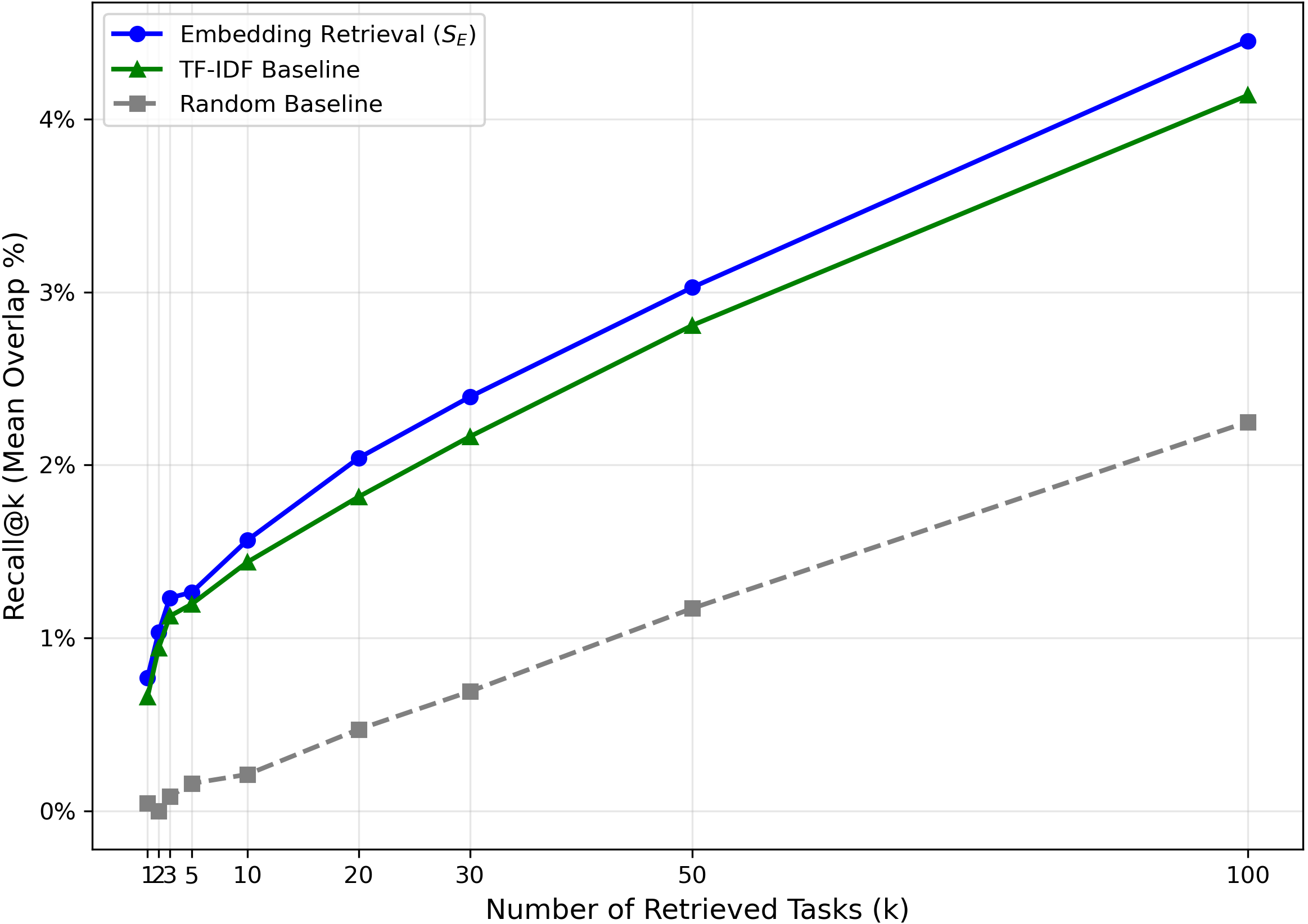}
    \vspace{-0.6em}
    \caption{
    {Oracle} retrieval consistency. {Embedding retrieval matches {oracle}-relevant tasks better than random retrieval and matches or exceeds TF-IDF.}
    }
    \label{fig:oracle-retrieval}
    \vspace{-0.6em} 
    \end{figure}


\begin{figure}[t]
\centering
\includegraphics[width=0.74\textwidth]{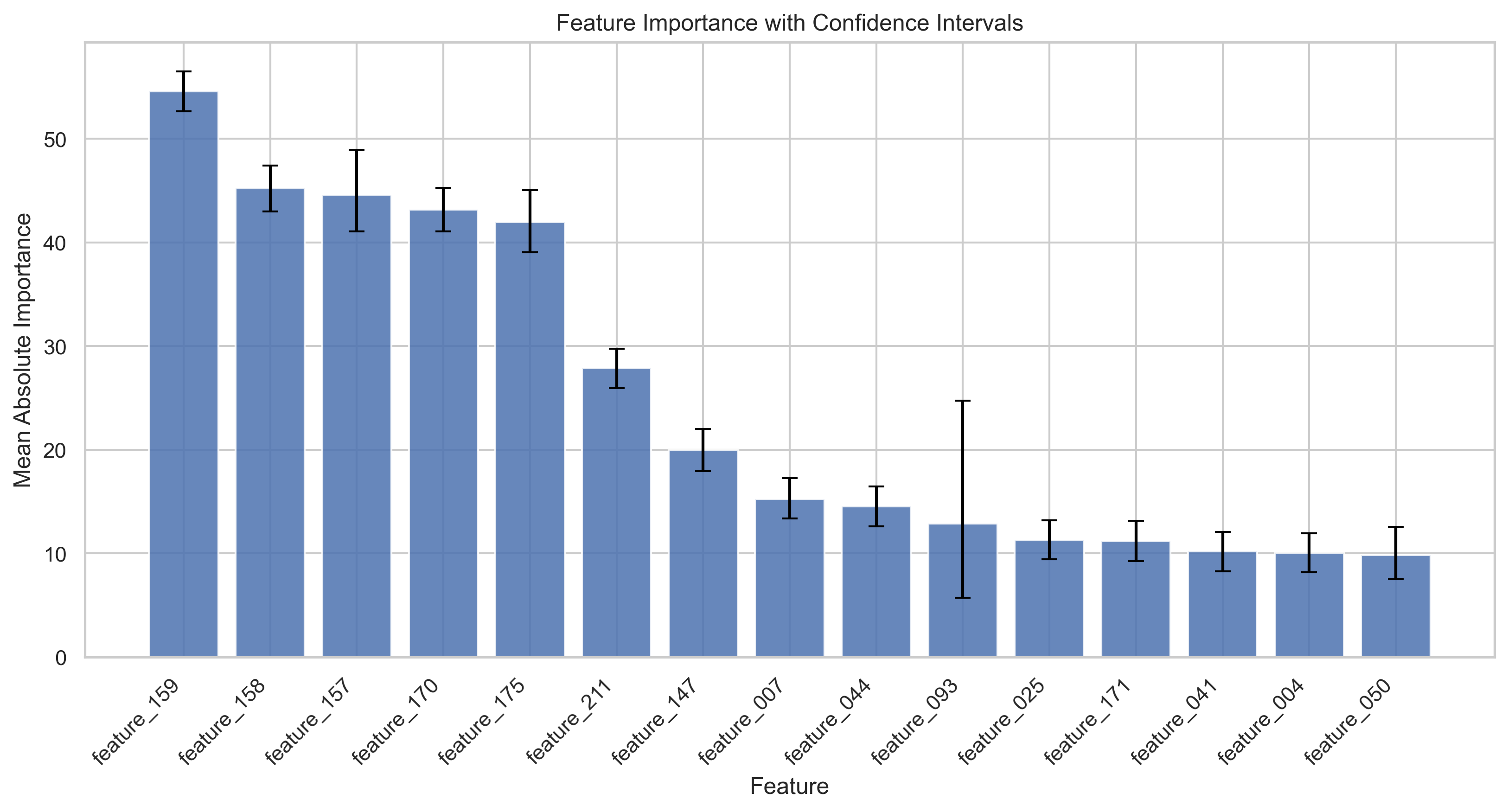}
\vspace{-0.6em}
\caption{
Feature-level interpretability with uncertainty. RAIL ranks diagnostic features by posterior coefficient magnitude and reports uncertainty intervals. 
}
\label{fig:feature-ci}
\vspace{-0.6em} 
\end{figure}


    \begin{figure}[!htb]
    \centering
    \begin{subfigure}{0.37\textwidth}
    \centering
    \includegraphics[width=\linewidth]{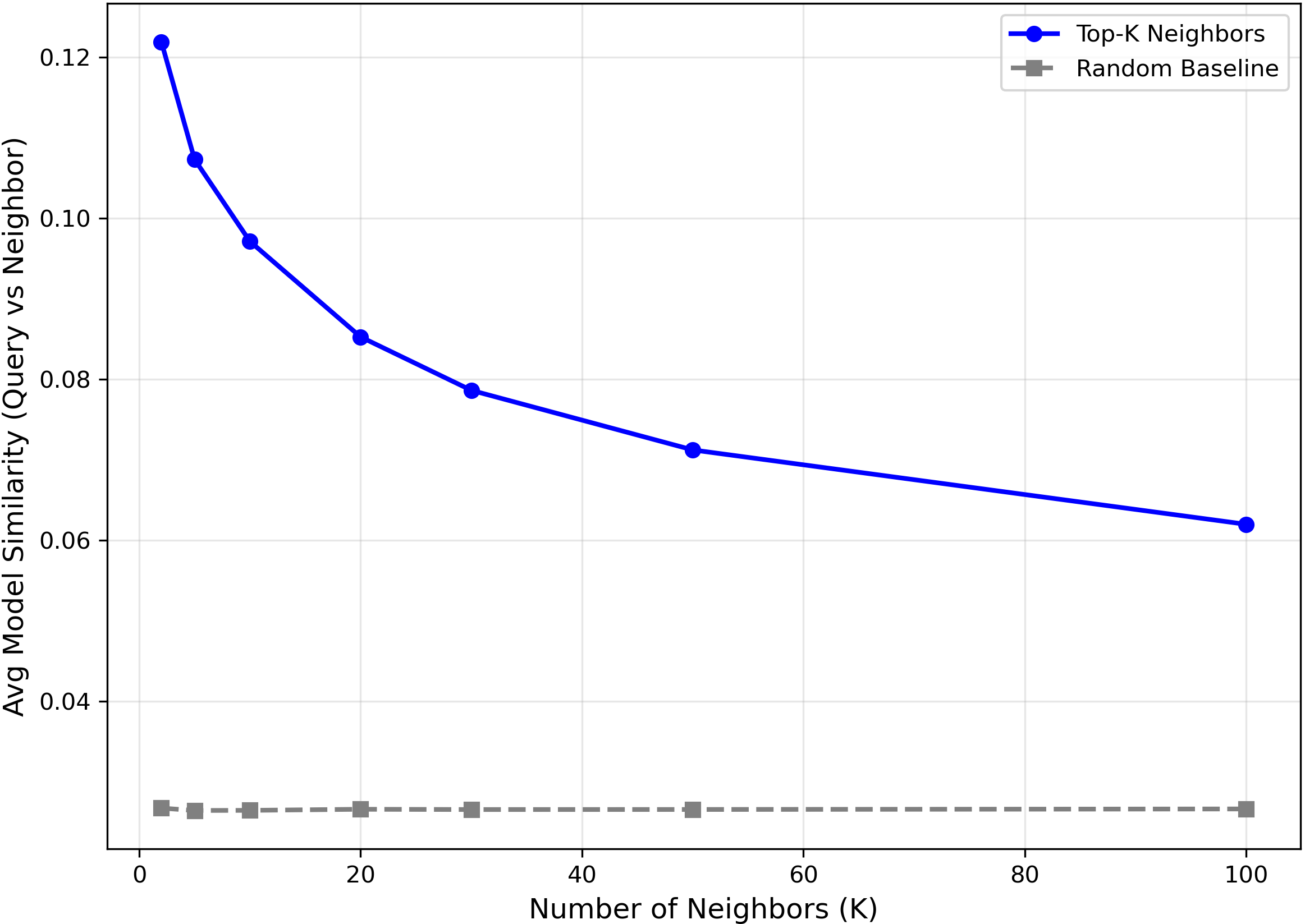}
    \caption{Query vs.\ retrieved neighbors.}
    \label{fig:query-neighbor-similarity}
    \end{subfigure}
    \hfill
    \begin{subfigure}{0.37\textwidth}
    \centering
    \includegraphics[width=\linewidth]{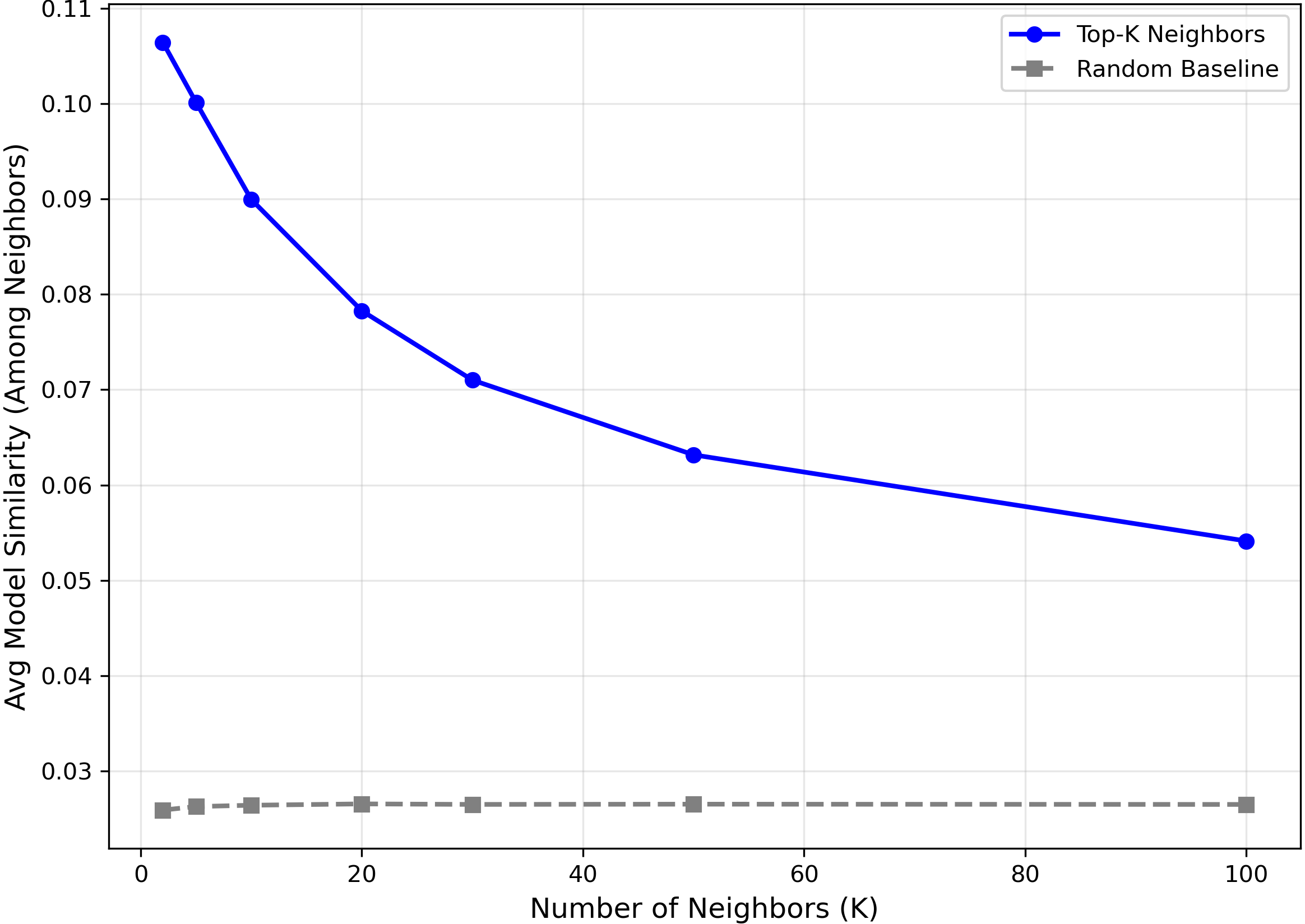}
    \caption{Pairwise similarity among neighbors.}
    \label{fig:neighbor-neighbor-similarity}
    \end{subfigure}
    \caption{
    Embedding--model alignment. Top-$k$ retrieved tasks are closer to the query and more coherent among themselves than random neighbors.
    }
    \label{fig:embedding-model-alignment}
    \vspace{-0.6em} 
    \end{figure}

\begin{figure}[!t] 
        \centering
        \includegraphics[width=0.76\textwidth]{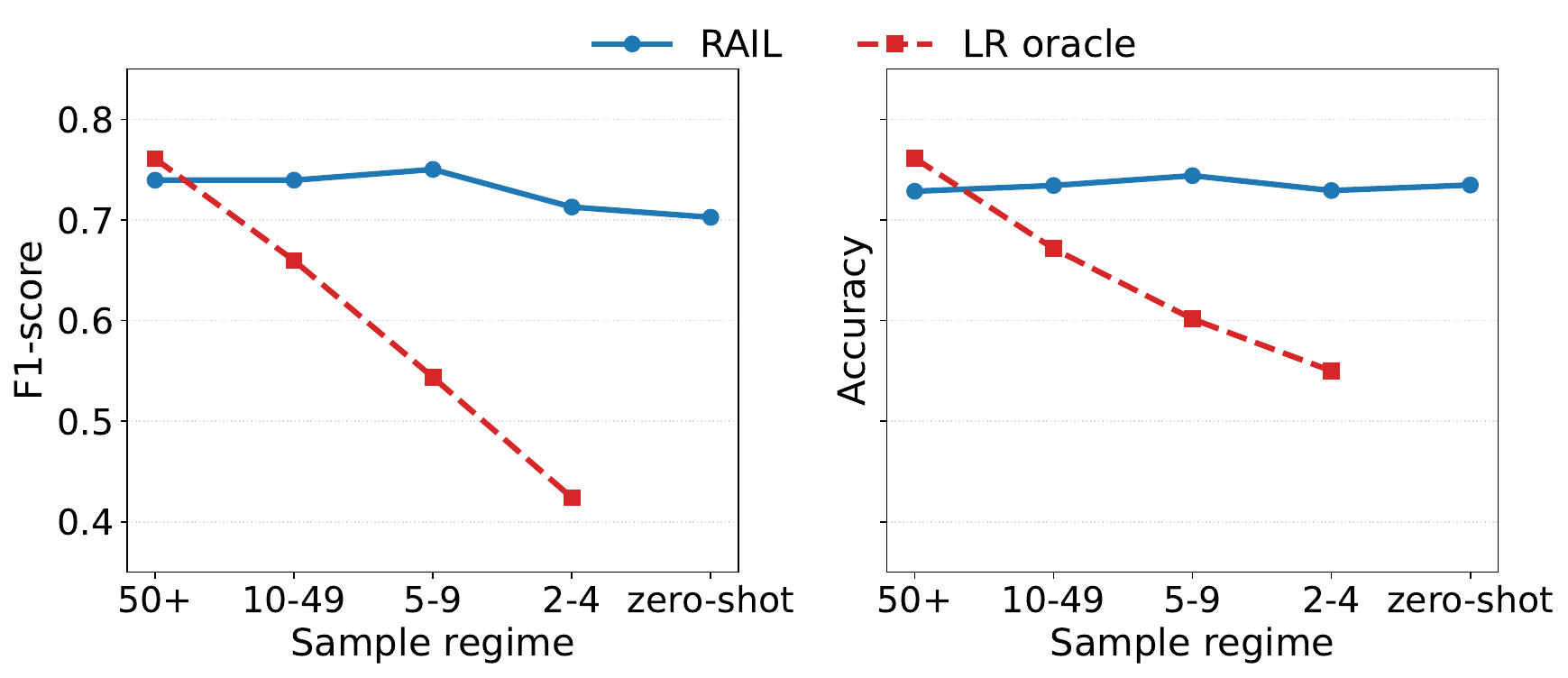}
        \caption{
        RAIL remains stable across sample-scarcity regimes, while {task-specific LR} degrades in the low-data tail. {In the legend, ``LR oracle'' denotes task-specific LR, which is trained on the target task's own labels.}
        }
        \label{fig:long-tail-performance}
        \vspace{-0.6em} 
        \end{figure}



\begingroup
\color{black}

\section{Additional Experimental Results}
\label{app:additional_experimental_results}

This appendix provides the complete empirical details supporting the results in the main paper. We organize the analyses by scientific question: full predictive performance and evaluation protocols; baseline and mechanism ablations; 
calibration and uncertainty; retrieval, memory, and overlap; generalization and robustness; extension beyond linear predictors; interpretability and computational scaling{; and LLM-based task-description expansion}. The patient- and era-disjoint benchmark is reported separately from the {sample-scarcity benchmark} throughout.



\subsection{Evaluation Protocols}
\label{app:full_predictive_results}

The {sample-scarcity benchmark} contains $4{,}412$ source-memory tasks and $3{,}075$ held-out zero-shot target tasks. The strict patient- and era-disjoint benchmark was rebuilt independently, with training data from 2008--2013 and evaluation data from 2014--2019. 
{The strict benchmark contains $4{,}395$ source-memory tasks and $3{,}092$ zero-shot target tasks. 
Memory and trainable RAIL modules are fit separately for these two benchmarks.
\subsection{Baselines and Mechanism Ablations}
\label{app:baselines_mechanism_ablations}

{Tab.~\ref{tab:baseline-ablation} in the main paper} reports the principal predictive and interpretability comparison. {Tab.}~\ref{tab:ablation-suite} summarizes the broader diagnostic suite and clarifies which component is tested by each analysis. These analyses separate the contribution of task-text representation, meaningful retrieval, learned coefficient synthesis, and posterior uncertainty. 


\begin{table}[!ht]
\centering
\caption{Summary of the baseline, ablation, and diagnostic suite. Each analysis isolates a distinct part of the retrieval-conditioned model-generation mechanism.}
\label{tab:ablation-suite}
\resizebox{0.99\columnwidth}{!}{%
\begin{tabular}{lll}
\toprule
Analysis group & Variants tested & Purpose \\
\midrule
Text encoder & DistilBERT, MedEmbed & Tests the task-description representation used for retrieval and synthesis. \\
Retrieval perturbation & Normal, random, distant & Tests whether semantically meaningful retrieval is necessary. \\
Transfer baseline & Top-$1$ LR, Top-$k$ coefficient average & Separates learned synthesis from direct model transfer. \\
Gradient-based adaptation & {MAML$_{\mathrm{LR}}$, MAML$_{\mathrm{MLP}}$} & Compares with gradient-based task adaptation. \\
Interpretable model class & {Task-specific LR and GAM}, {RAIL$_{\mathrm{GAM}}$} & Tests whether RAIL extends beyond linear predictors. \\
{Optional gate (Eq.~\ref{eq:posterior-mean-gate})} & Disabled, learned global gate & {Tests whether interpolating with a task-specific estimator improves few-shot adaptation.} \\
Embedding corruption & Gaussian noise & Tests sensitivity to semantic corruption. \\
Embedding compression & PCA-32, PCA-64 & Tests whether retrieval structure is compressible. \\
Embedding randomization & Random, shuffled & Tests whether performance exceeds chance structure. \\
Retrieval diagnostics & Query-neighbor, neighbor-neighbor, {Oracle} Recall@$k$ & Tests embedding--model alignment. \\
Uncertainty diagnostics & Retrieval, parameter, predictive uncertainty & Tests reliability and failure detection. \\
\bottomrule
\end{tabular}%
}
\vspace{-0.4em} 
\end{table}

{The optional gate $\alpha$ in Eq.~\ref{eq:posterior-mean-gate} is a single global per-coordinate gate over the coefficient representation; with the gate disabled, the posterior mean equals the retrieval-conditioned mean, 
and the main results use the enabled gate. 
The learned gate is prior-leaning: its mean is $0.705$ (standard deviation $0.241$), and $69\%$ of coordinates exceed $0.5$. Enabling it changes AUROC by $+0.003$, AUPRC by $+0.003$ to $+0.004$, and accuracy by at most $0.002$ in every regime (Tab.~\ref{tab:gate-ablation}), so the two settings perform essentially the same. This is expected: in both settings, the target's own task-specific LR, fitted on its training examples, can be retrieved from the parametric memory at few-shot inference, and the target-task data already supervise the generator during training, so the explicit estimator $\hat\theta^{\mathrm{self}}_t$ adds little new information for few-shot settings, where number of samples to train $\hat\theta^{\mathrm{self}}_t$ is already extremely scarce.}

\begin{table}[!ht]
\centering
\caption{{Effect of the optional gate (Eq.~\ref{eq:posterior-mean-gate}). The main results use the enabled gate.}}
\label{tab:gate-ablation}
\begin{tabular}{lcccccc}
\toprule
& \multicolumn{2}{c}{Accuracy} & \multicolumn{2}{c}{AUROC} & \multicolumn{2}{c}{AUPRC} \\
Regime & Disabled & Enabled & Disabled & Enabled & Disabled & Enabled \\
\midrule
$10$--$49$ & 0.734 & 0.736 & 0.796 & 0.799 & 0.746 & 0.749 \\
$5$--$9$   & 0.744 & 0.746 & 0.810 & 0.813 & 0.762 & 0.765 \\
$2$--$4$   & 0.732 & 0.732 & 0.790 & 0.793 & 0.736 & 0.740 \\
\bottomrule
\end{tabular}
\vspace{-0.4em} 
\end{table}

The gradient-based references provide a complementary comparison. {With the same $2$--$4$ training examples per task}, {MAML$_{\mathrm{LR}}$} obtains accuracy $0.653$ and AUROC $0.702$, compared with RAIL's {$0.732$} and {$0.793$}. A non-interpretable {MAML$_{\mathrm{MLP}}$} obtains accuracy $0.684$ and AUROC $0.754$; it is therefore useful as an out-of-class reference while not providing the feature-level interpretability or coefficient uncertainty of RAIL.

\subsection{Calibration, Uncertainty, and Selective Prediction}
\label{app:calibration_uncertainty}

The main paper reports pooled AUROC, AUPRC, ECE, MCE, and Brier score for {the $5$--$9$, $2$--$4$, and zero-shot regimes; the $50+$ split is used to fit the temperature}. AUROC and AUPRC are computed from pooled predictions within each regime. ECE uses $15$ equal-width probability bins, MCE is the maximum bin-wise calibration gap, and Brier score is the mean squared error of the predicted probability. For the reported calibration results, {$100$} Monte Carlo coefficient samples are averaged, one temperature is fitted on the $50+$ evaluation split, and that temperature is transferred unchanged to the lower-resource and zero-shot regimes. {The task-specific LR baseline is temperature-scaled with the same protocol, using a temperature fitted on the $50+$ split.} 

\begin{table}[!ht]
\centering
\caption{Tied- versus untied-variance posterior comparison. {The untied model is trained separately and evaluated with the same protocol.} 
}
\label{tab:untied-variance-results}
\begin{tabular}{lcccc}
\toprule
Metric & Tied few-shot & Untied few-shot & Tied zero-shot & Untied zero-shot \\
\midrule
Accuracy & 0.732 & 0.738 & 0.734 & 0.742 \\
AUROC   & {0.793} & 0.796 & 0.793 & 0.799 \\
AUPRC   & {0.740} & 0.751 & 0.739 & 0.757 \\
ECE     & 0.068 & 0.066 & 0.067 & 0.069 \\
MCE     & 0.105 & 0.128 & 0.111 & 0.100 \\
Brier   & 0.193 & 0.182 & 0.194 & 0.181 \\
\bottomrule
\end{tabular}
\vspace{-0.4em} 
\end{table}

\subsection{Untied Posterior Variance} \label{app:untied_posterior}
To relax the tied-variance assumption $\sigma^2_{q,t}\equiv\sigma^2_{\phi,t}$, we untie the posterior by precision addition. For each coefficient coordinate $r$,
\begin{align}
\frac{1}{\sigma^{2\,(r)}_{q,t}}
=
\frac{1}{\sigma^{2\,(r)}_{\phi,t}}
+
n_t\,\Delta^{(r)}_{t},
\qquad
\Delta^{(r)}_{t}
=
\mathrm{softplus}\big(h_\psi(\hat\theta^{\mathrm{self}}_t)\big)^{(r)}
\ge 0,
\label{eq:untied-precision}
\end{align}
where $n_t=|\mathcal D_t|$ is the number of target-task training examples and $h_\psi$ is a learned map from the task-specific estimator to a per-coordinate precision rate. The posterior variance is then
$\sigma^{2\,(r)}_{q,t}=\sigma^{2\,(r)}_{\phi,t}\big/\big(1+n_t\,\Delta^{(r)}_{t}\,\sigma^{2\,(r)}_{\phi,t}\big)$, which contracts as target-task evidence accumulates and recovers the prior variance when $n_t=0$, so zero-shot inference is unchanged. Because the variances are no longer shared, we replace the mean-discrepancy coefficient KL with the general two-Gaussian KL in Eq.~\ref{eq:kl-theta-general} and retrain the model from scratch. To quantify the contraction, we take $300$ supervised tasks, read each task's prior variance and learned precision rate from a forward pass, evaluate $\sigma_{q,t}/\sigma_{\phi,t}$ at increasing $n_t$, and average over coordinates and then over tasks.

The untied posterior improves discrimination and Brier score while changing ECE only slightly (Appendix Tab.~\ref{tab:untied-variance-results}). Its posterior standard-deviation ratio relative to the prior scale is $1.000$, $0.974$, $0.898$, $0.829$, and $0.575$ {with $0$, $100$, $500$, $1000$, and $5000$} target-task examples, respectively. The untied model requires a temperature of approximately $3.55$, compared with approximately $5.15$ for the tied model under the corresponding calibration procedure. These results support the tied formulation as a stable predictive simplification while showing that the framework can accommodate more expressive uncertainty when needed.

\begin{table}[!ht]
\centering
\caption{Uncertainty signals used in the RAIL reliability analyses.}
\label{tab:uncertainty-definitions}
\begin{tabular}{lll}
\toprule
Signal & Definition & Level \\
\midrule
Retrieval uncertainty & $-\sum_j q_\psi(I_t=j)\log q_\psi(I_t=j)$ & Task \\
Parameter uncertainty & $\sum_r \sigma^2_{q,t,r}$ & Task/model \\
Predictive uncertainty & $-\bar p\log\bar p-(1-\bar p)\log(1-\bar p)$ & Sample \\
Confidence & $\max(\bar p,1-\bar p)$ & Sample \\
\bottomrule
\end{tabular}
\vspace{-0.4em} 
\end{table}

\begin{figure}[!htb]
\centering
\begin{subfigure}{0.48\linewidth}
\centering
\includegraphics[width=\linewidth]{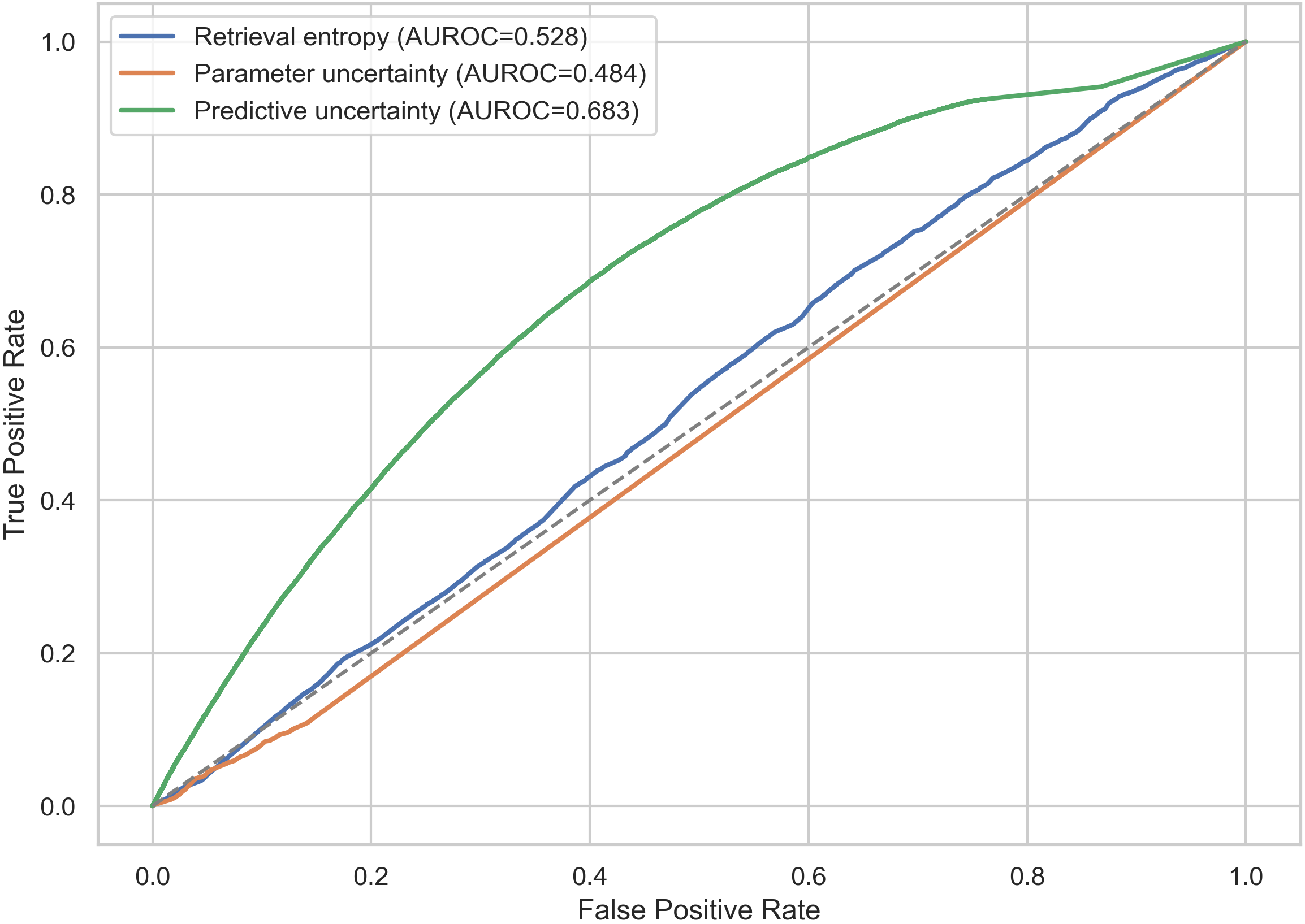}
\caption{Failure-detection ROC.}
\end{subfigure}
\hfill
\begin{subfigure}{0.48\linewidth}
\centering
\includegraphics[width=\linewidth]{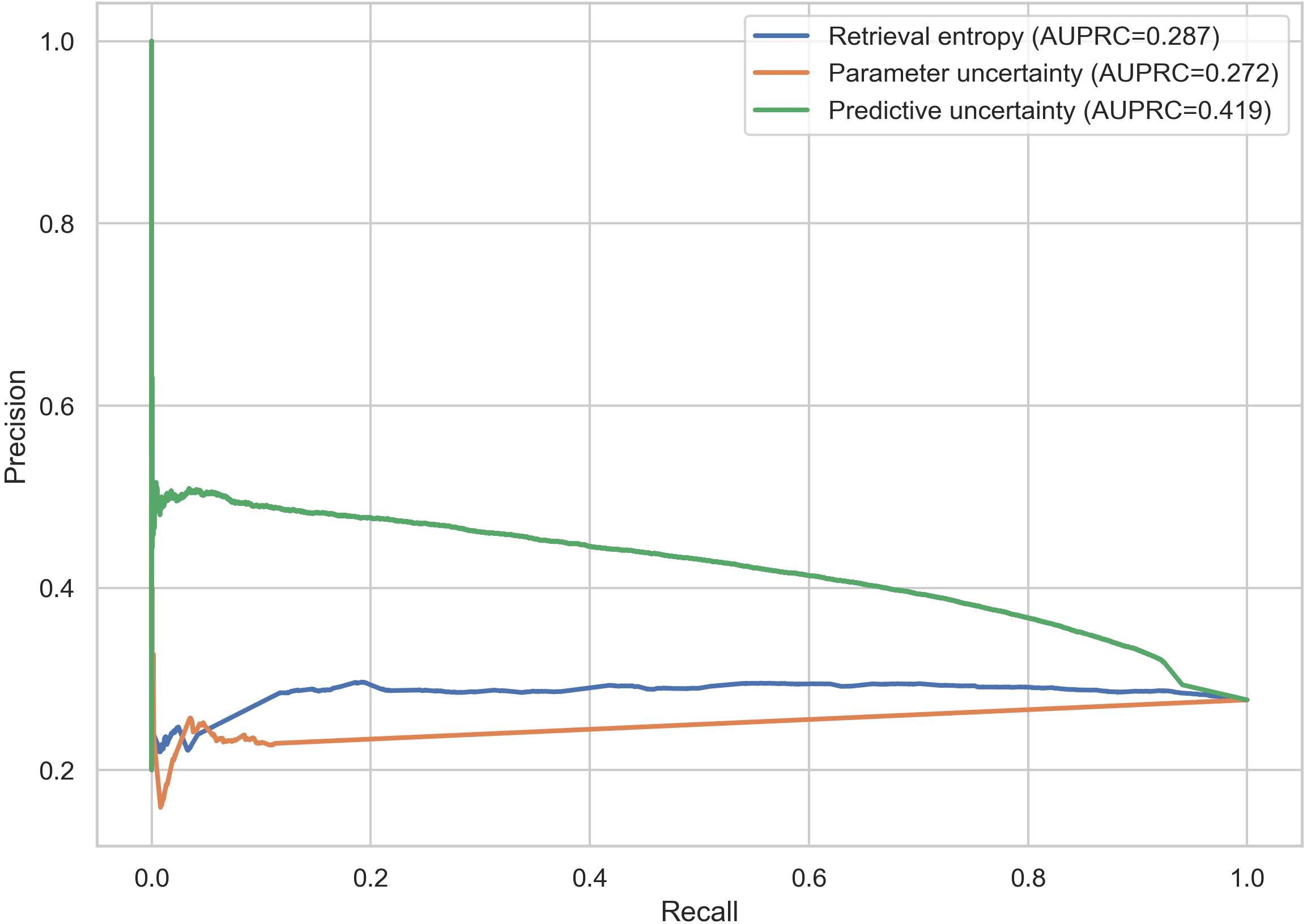}
\caption{Failure-detection precision--recall curve.}
\end{subfigure}
\caption{Failure detection using uncertainty. Predictive uncertainty achieves AUROC $0.683$ and AUPRC $0.419$ for identifying incorrect predictions and outperforms retrieval entropy and parameter uncertainty in the corresponding analyses.}
\label{fig:app-failure-detection}
\vspace{-0.6em} 
\end{figure}

\begin{figure}[!htb]
        \vspace{-1.0em}
        \centering
        \begin{subfigure}{0.38\textwidth}
        \centering
        \includegraphics[width=\linewidth]{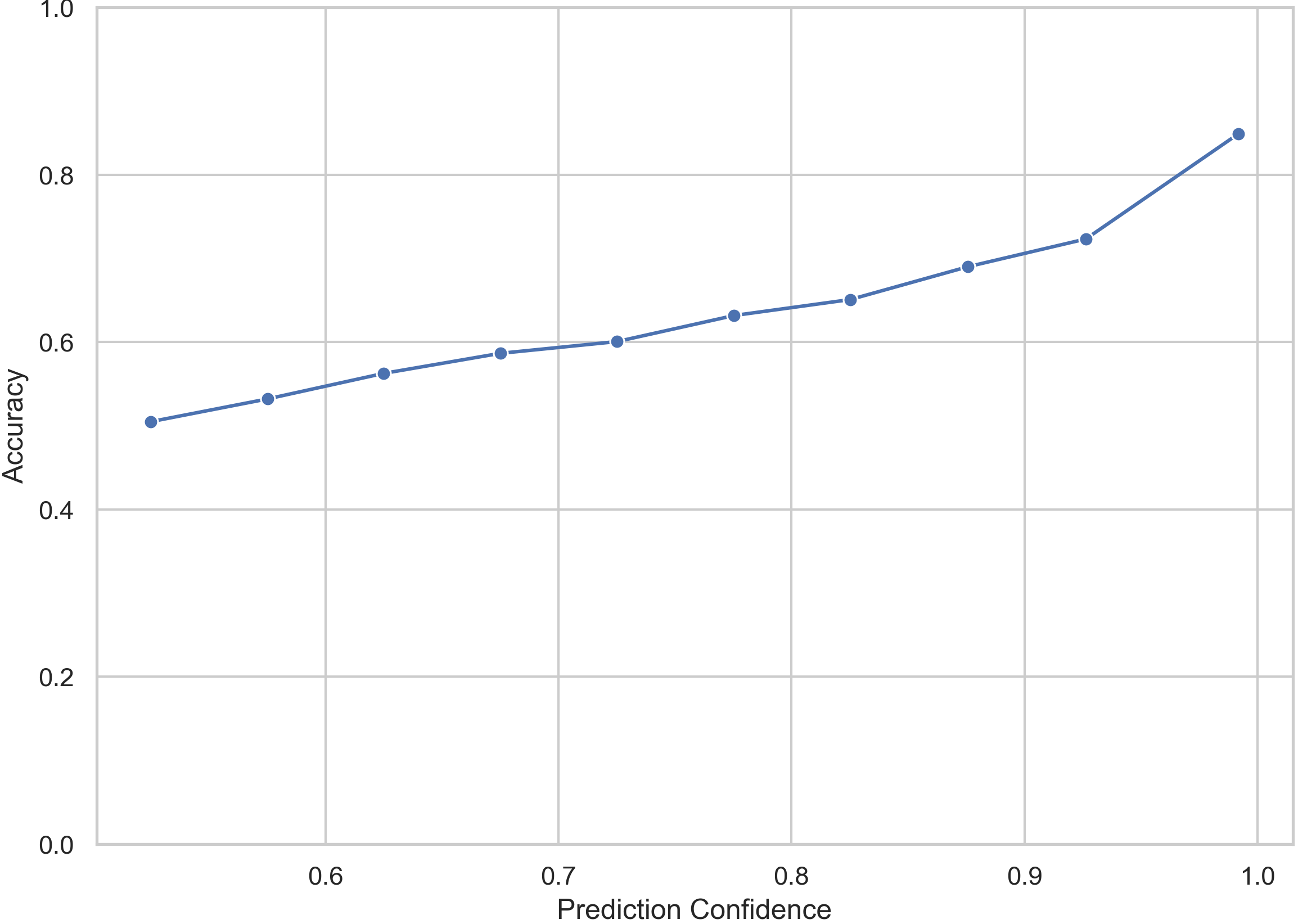}
        \vspace{-0.6em}
        \caption{Confidence vs.\ accuracy.}
        \label{fig:confidence-accuracy}
        \end{subfigure}
        \hfill
        \begin{subfigure}{0.38\textwidth}
        \centering
        \includegraphics[width=\linewidth]{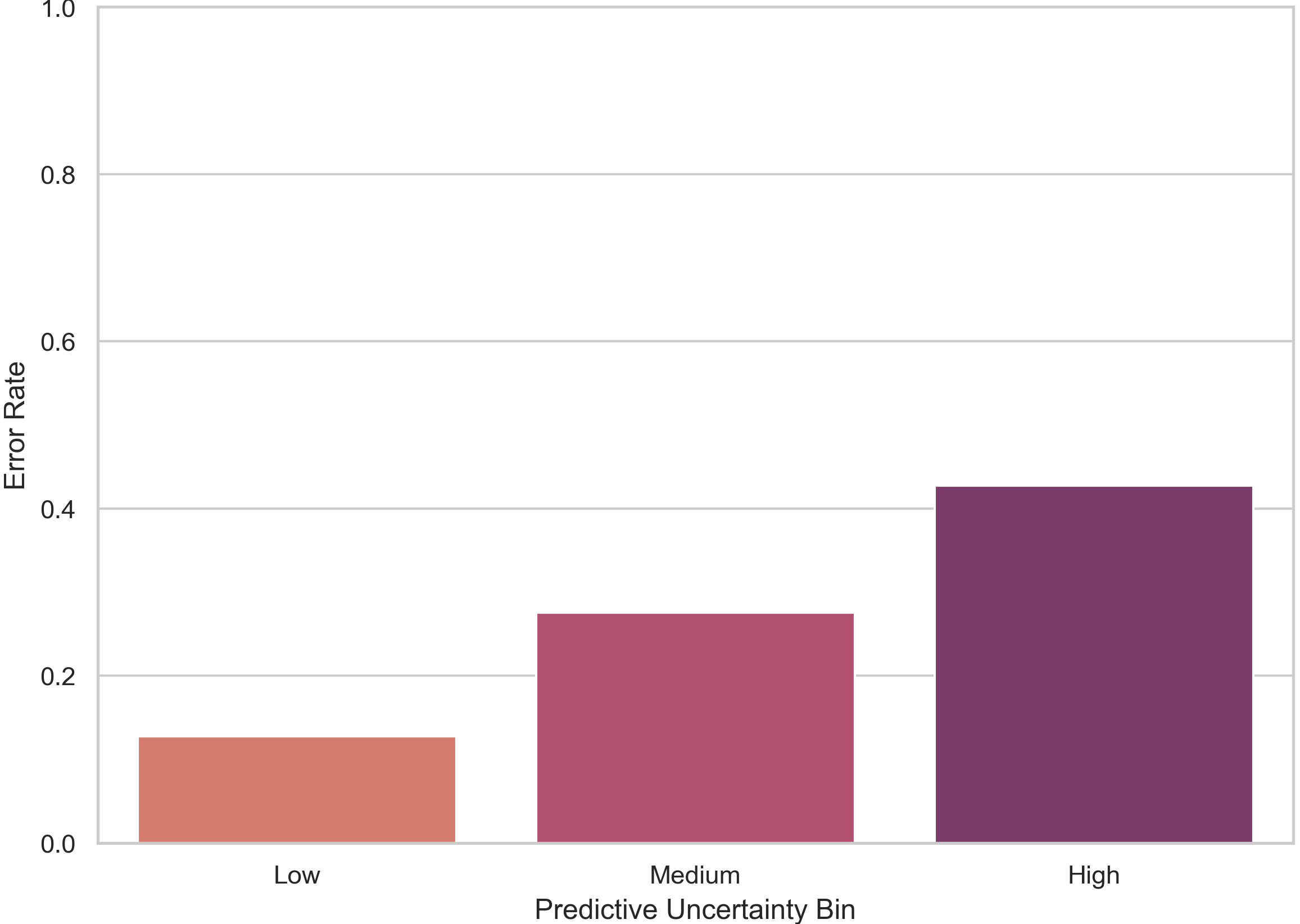}
        \vspace{-0.6em}
        \caption{Error vs.\ uncertainty.}
        \label{fig:error-uncertainty}
        \end{subfigure}
        \vspace{-0.5em}
        \caption{
        Predictive uncertainty diagnostics. Confidence tracks empirical accuracy, while higher uncertainty corresponds to higher error.
        }
        \label{fig:uncertainty-basic}
        \vspace{-0.6em} 
        \end{figure}

The uncertainty signals operate at different levels: retrieval entropy measures ambiguity in the source-task distribution, parameter uncertainty measures uncertainty in the synthesized coefficient vector, and predictive entropy measures uncertainty in the final sample-level probability. Predictive uncertainty is most directly aligned with prediction failure, whereas retrieval and parameter uncertainty provide complementary information about the source of model unreliability. The selective-prediction analysis in the main paper uses these scores to retain lower-risk predictions at reduced coverage.

\subsection{Retrieval, Memory, and Overlap Analysis}
\label{app:retrieval_overlap_memory}

We evaluate whether RAIL depends on meaningful organization in task-embedding space or on a small number of near-duplicate source tasks. The representation perturbation experiments include Gaussian noise, embedding compression, random embeddings, and shuffled embeddings. They show that noise degrades performance, compact embeddings preserve useful retrieval structure, and random or shuffled representations substantially reduce retrieval quality.

\begin{figure}[!htb]
\centering
\begin{subfigure}{0.32\linewidth}
\centering
\includegraphics[width=\linewidth]{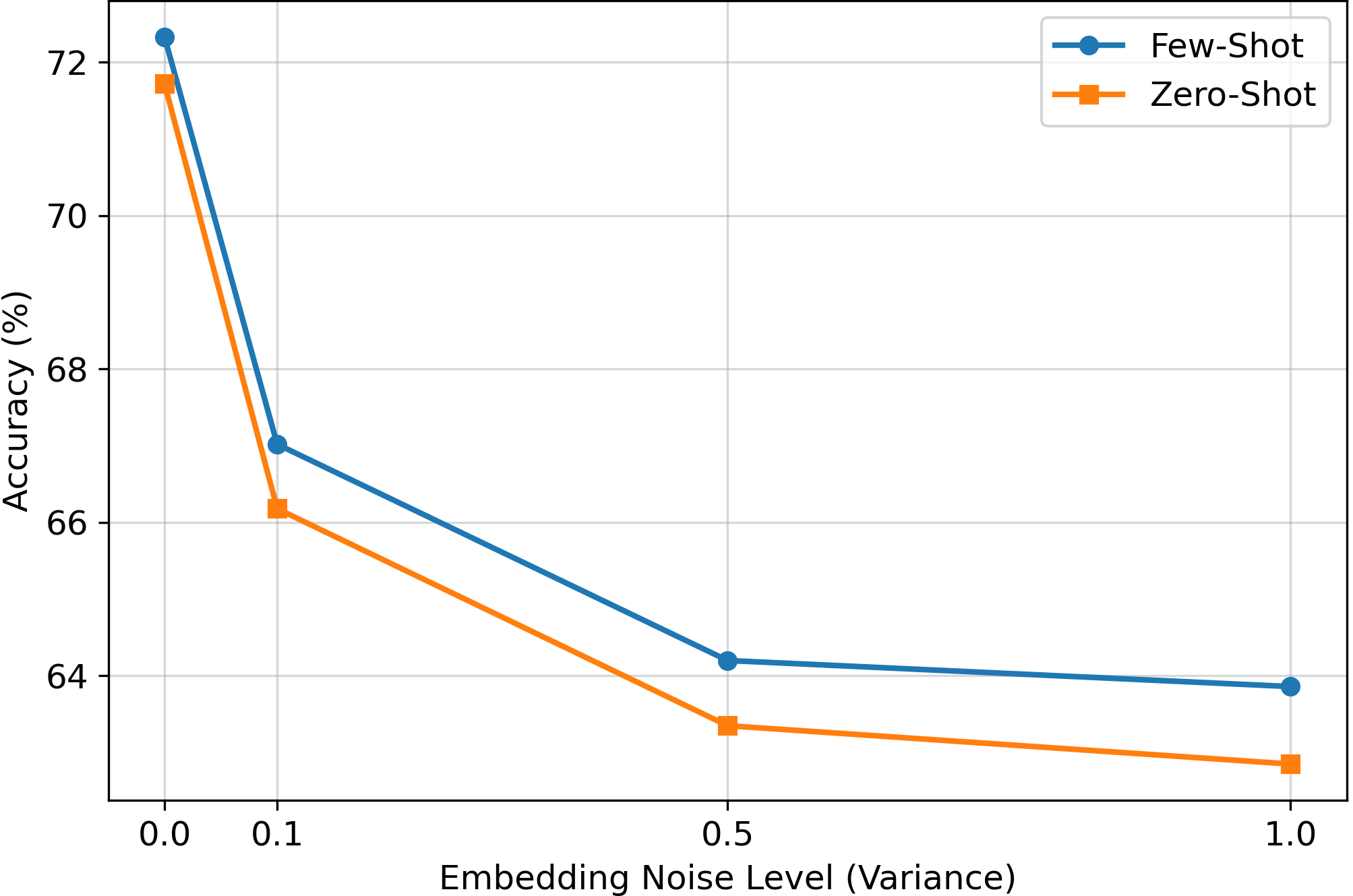}
\caption{Noise perturbation.}
\end{subfigure}
\hfill
\begin{subfigure}{0.32\linewidth}
\centering
\includegraphics[width=\linewidth]{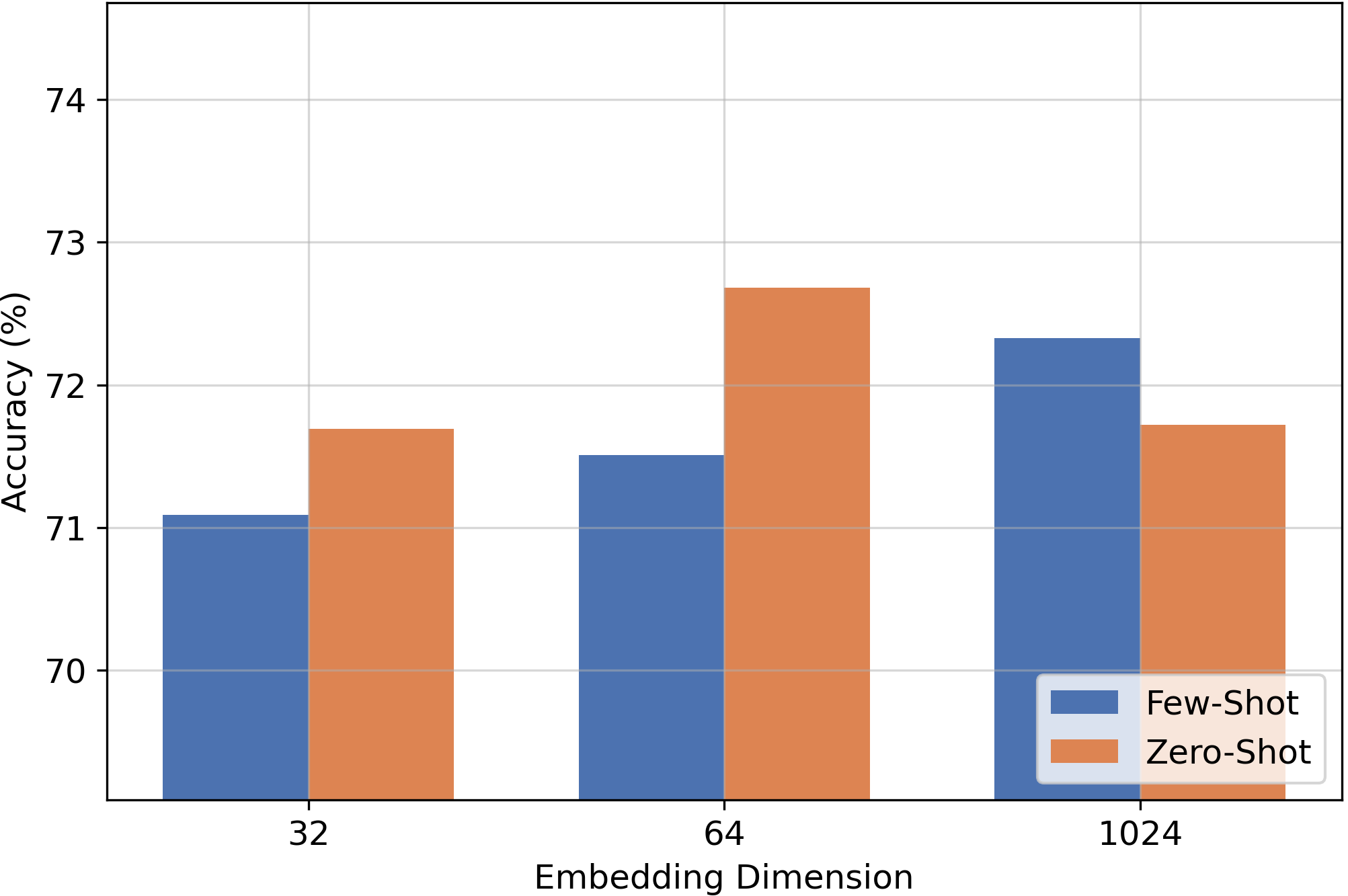}
\caption{Embedding dimension.}
\end{subfigure}
\hfill
\begin{subfigure}{0.32\linewidth}
\centering
\includegraphics[width=\linewidth]{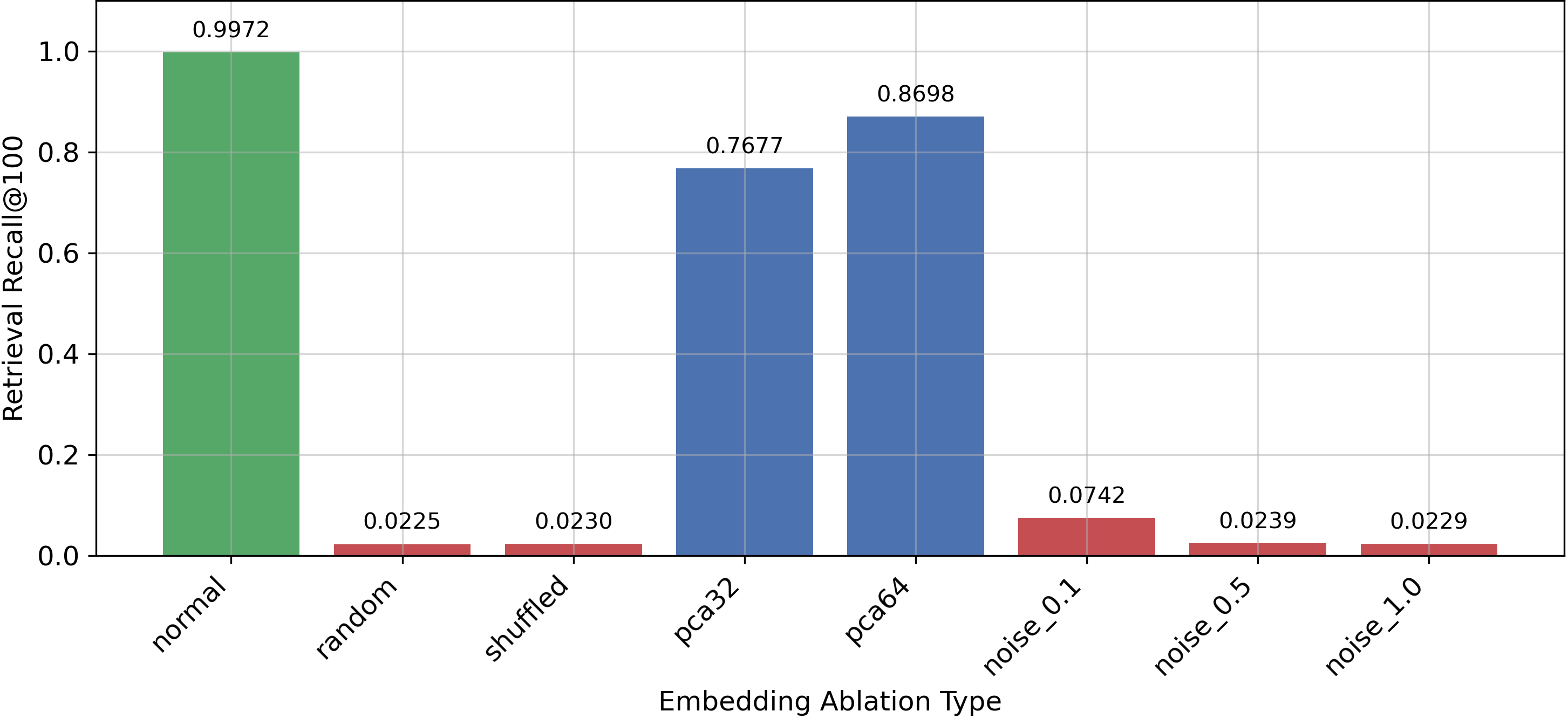}
\caption{{Retrieval recall under embedding ablations.}}
\end{subfigure}
\caption{Additional task-representation ablations. Semantic corruption degrades performance, while compressed representations retain useful retrieval structure and random or shuffled representations collapse retrieval quality. 
}
\label{fig:app-embedding-ablation}
\vspace{-0.6em} 
\end{figure}

The retrieval diagnostics further compare query--neighbor similarity, neighbor--neighbor coherence, and {oracle} overlap. Retrieved neighbors are closer to the query and more coherent among themselves than random neighbors. Embedding retrieval also achieves higher {oracle} overlap across retrieval depths than random retrieval and {matches or exceeds TF-IDF (Figs.~\ref{fig:embedding-model-alignment} and~\ref{fig:oracle-retrieval})}, indicating that the learned task representation captures similarity relevant for model transfer rather than only lexical overlap.

\begin{table}[!ht]
\centering
\caption{Zero-shot accuracy as a function of memory size. {Smaller memories: mean $\pm$ standard deviation over $10$ random subsets of the full memory; full memory size = $4{,}412$.
}}
\label{tab:memory-size-ablation}
\begin{tabular}{lccccc}
\toprule
Memory size & $100$ & $500$ & $1{,}000$ & $2{,}000$ & $4{,}412$ \\
\midrule
Zero-shot accuracy & $0.704\!\pm\!0.007$ & $0.721\!\pm\!0.003$ & $0.728\!\pm\!0.002$ & $0.731\!\pm\!0.002$ & {$0.734$} \\
\bottomrule
\end{tabular}
\vspace{-0.4em} 
\end{table}

The memory-size analysis uses ten random memory subsets at the smaller sizes. A memory of $100$ tasks is close to the retrieval-free reference, whereas a memory of approximately $1{,}000$ tasks recovers most of the full-memory performance. This pattern indicates that memory coverage, rather than only the existence of a retrieval module, determines the quality of the synthesized prior.

\begin{table}[!ht]
\centering
\caption{Near-neighbor and overlap controls on the strict patient- and era-disjoint benchmark. AUROC and AUPRC are reported for zero-shot targets. {Random-removal entries are mean $\pm$ standard deviation over {$10$} random draws of the same size.}}
\label{tab:near-neighbor-overlap-controls}
\resizebox{0.80\columnwidth}{!}{%
\begin{tabular}{llcc}
\toprule
Control family & Memory operation & AUROC & AUPRC \\
\midrule
MedEmbed & No removal & 0.7552 & 0.7267 \\
MedEmbed & Remove nearest $10$ & 0.7530 & 0.7252 \\
MedEmbed & Remove nearest $100$ & 0.7440 & 0.7187 \\
MedEmbed & Remove nearest $2{,}000$ & 0.7038 & 0.6811 \\
MedEmbed & {Random removal}, $100$ & $0.7550\!\pm\!0.0002$ & $0.7269\!\pm\!0.0004$ \\
MedEmbed & {Random removal}, $2{,}000$ & $0.7507\!\pm\!0.0009$ & $0.7231\!\pm\!0.0013$ \\
TF-IDF & Remove nearest $100$ & 0.7463 & 0.7198 \\
TF-IDF & {Random removal}, $100$ & $0.7550\!\pm\!0.0004$ & $0.7272\!\pm\!0.0007$ \\
ICD-10-PCS & Remove tasks sharing first $3$ characters & 0.7519 & 0.7229 \\
ICD-10-PCS & {Random removal, same number} & $0.7551\!\pm\!0.0001$ & $0.7267\!\pm\!0.0003$ \\
ICD-10-PCS & Remove tasks sharing first $4$ characters & 0.7550 & 0.7266 \\
ICD-10-PCS & {Random removal, same number} & $0.7551\!\pm\!0.0001$ & $0.7266\!\pm\!0.0001$ \\
\bottomrule
\end{tabular}%
}
\vspace{-0.4em} 
\end{table}

On the strict split, there are no exact normalized-description matches between the zero-shot targets and the memory tasks ($0/3{,}092$). At least one memory task shares the first three ICD-10-PCS characters with $93.8\%$ of targets and the first four characters with $51.1\%$ of targets; the nearest memory task shares an average of $2.98$ leading characters. Removing the complete initial candidate set of $100$ MedEmbed neighbors causes a modest AUROC decrease from $0.7552$ to $0.7440$, whereas removing a much broader nearest neighborhood is substantially more damaging. {Random removals of the same size} remain close to the {full-memory result}, indicating that the effect is associated with semantic coverage rather than simply removing the same number of memory entries.

\subsection{Generalization and Robustness}
\label{app:generalization_robustness}

The strict benchmark separates both patients and calendar eras. RAIL is retrained on the 2008--2013 cohort, its parametric memory is refit using only the corresponding source tasks, and evaluation is performed on the 2014--2019 cohort. Results are in Tab.~\ref{tab:strict-split-generalization} of the main paper. 


\paragraph{Partial feature overlap.}
To test sensitivity to missing inputs within the shared diagnostic space, we mask $0\%$, $20\%$, $50\%$, $80\%$, and $100\%$ of the $217$ shared features. The zero-shot AUROC values are $0.793$, $0.738$, $0.666$, $0.603$, and $0.500$, respectively. The smooth degradation indicates that the learned predictor uses distributed information across the shared feature space.

\begin{table}[!ht]
\centering
\caption{Zero-shot AUROC under partial masking of the shared diagnostic-feature space.}
\label{tab:feature-masking-results}
\begin{tabular}{lccccc}
\toprule
Masked features & $0\%$ & $20\%$ & $50\%$ & $80\%$ & $100\%$ \\
\midrule
Zero-shot AUROC & 0.793 & 0.738 & 0.666 & 0.603 & 0.500 \\
\bottomrule
\end{tabular}
\vspace{-0.4em} 
\end{table}


\paragraph{Description expansion.}
Expanding the procedure descriptions improves retrieval and synthesis modestly without changing the model class. The full comparison is given in {Tab.}~\ref{tab:description-expansion-results}.

\begin{table}[!ht]
\centering
\caption{Effect of expanding the natural-language task descriptions using {\texttt{gpt-4.1-mini}} (Appendix~\ref{app:description-expansion}). 
}
\label{tab:description-expansion-results}
\begin{tabular}{llcccc}
\toprule
Regime & Description & Accuracy & F1 & AUROC & AUPRC \\
\midrule
Few-shot & Default & {0.732} & 0.745 & {0.793} & {0.740} \\
Few-shot & Expanded & 0.739 & 0.765 & 0.797 & 0.745 \\
Zero-shot & Default & {0.734} & 0.747 & 0.793 & 0.739 \\
Zero-shot & Expanded & 0.742 & 0.767 & 0.802 & 0.750 \\
\bottomrule
\end{tabular}
\vspace{-0.4em} 
\end{table}

\paragraph{Prevalence sensitivity.}
The {sample-scarcity benchmark} uses balanced binary task construction. To assess the effect of changing the positive class prevalence, {we reweight the zero-shot negatives so that the positive class has prevalence $\pi\in\{0.50,0.20,0.10,0.05,0.01\}$ and apply a Bayes base-rate correction to the temperature-scaled predictions.}

\begin{table}[!ht]
\centering
\caption{Zero-shot AUPRC under prevalence reweighting. Lift is measured relative to the corresponding positive-class prevalence.}
\label{tab:prevalence-reweighting}
\begin{tabular}{lcc} 
\toprule
Positive prevalence & AUPRC & Lift over prevalence \\
\midrule
0.50 & 0.739 & $1.5\times$ \\
0.20 & 0.437 & $2.2\times$ \\
0.10 & 0.251 & $2.5\times$ \\
0.05 & 0.153 & $3.1\times$ \\
0.01 & 0.032 & $3.2\times$ \\
\bottomrule
\end{tabular}
\vspace{-0.4em} 
\end{table}


The decrease in AUPRC follows the changing base rate, while the lift remains above one across all tested prevalences. {The results therefore characterize sensitivity to prior class balance.}

\subsection{\texorpdfstring{Generality Beyond Linear Models: {RAIL$_{\mathrm{GAM}}$}}{Generality Beyond Linear Models: RAIL-GAM}} 
\label{app:gam_extension}

RAIL's retrieval-to-model-generation mechanism is not tied to logistic regression. It requires a shared parameterization in which retrieved task information can be synthesized into an interpretable target-task predictor. To demonstrate this more general formulation, we replace the linear feature map {$\Phi(x)=(1,x)$} with a fixed cubic B-spline basis while retaining the same task encoder, retrieval mechanism, coefficient synthesizer, posterior uncertainty, and prediction procedure.

The GAM basis uses $4$ spline coefficients per diagnostic feature and one intercept, giving ${d=}217\times 4+1=869$ synthesized coefficients. Quantile-based knots are fixed using the global training feature matrix, so the basis is shared across tasks and can be used for zero-shot synthesis. The resulting predictor exposes a shape function for each diagnostic feature, while the posterior over synthesized coefficients provides uncertainty for those shape functions.

\begin{table}[!ht]
\centering
\caption{Implementation details for the interpretable model-class extension. {RAIL$_{\mathrm{GAM}}$} changes the shared predictor basis while preserving the retrieval-conditioned synthesis and uncertainty framework.}
\label{tab:gam-implementation}
\begin{tabular}{lll}
\toprule
Component & {RAIL$_{\mathrm{LR}}$} & {RAIL$_{\mathrm{GAM}}$} \\ 
\midrule
Predictor basis & Original diagnostic features & Fixed cubic B-spline basis \\
Basis construction & {$\Phi(x)=(1,x)$} & Quantile knots from training features \\
Coefficient dimension {$d$} & {$218=217+1$} & $869=217\times4+1$ \\
Retrieval mechanism & Shared & Unchanged \\
Coefficient synthesis & Probabilistic & Unchanged \\
Uncertainty propagation & Posterior coefficients & Posterior basis coefficients \\
Zero-shot operation & Yes & Yes \\
\bottomrule
\end{tabular}
\vspace{-0.4em} 
\end{table}

The full model-class comparison is reported in {Tabs.~\ref{tab:baseline-ablation} and~\ref{tab:interpretable-model-classes}} in the main paper. {RAIL$_{\mathrm{GAM}}$ nearly matches RAIL$_{\mathrm{LR}}$ in few-shot accuracy while improving pooled AUROC and AUPRC in both few-shot and zero-shot evaluation.} The per-task GAM is weaker in the low-data regime and cannot operate zero-shot because it has no mechanism for synthesizing a target-task predictor without target labels. These results support RAIL as a general framework for retrieval-conditioned synthesis of interpretable predictors, rather than as a method specific to linear models.

\subsection{Interpretability Diagnostics and Computational Scaling}
\label{app:interpretability_efficiency}

RAIL exposes several complementary outputs for inspecting predictions and the source of model uncertainty. Generated coefficients provide feature-level diagnostic weights, posterior coefficient variance indicates uncertainty in individual feature contributions, top-$k$ feature frequency measures stability across posterior samples, retrieval posterior weights attribute the synthesized model to source tasks, and predictive confidence summarizes sample-level reliability.

\begin{table}[!ht]
\centering
\caption{Interpretability outputs produced by RAIL.}
\label{tab:interpretability-outputs}
\resizebox{0.99\columnwidth}{!}{%
\begin{tabular}{lll}
\toprule
Output & Level & Interpretation \\
\midrule
Generated coefficients $\theta_t$ & Task & Interpretable diagnostic-feature weights. \\
Posterior coefficient variance & Feature & Uncertainty in each feature's contribution. \\
Top-$k$ feature frequency & Feature & Stability of feature importance across posterior samples. \\
Retrieval posterior weights & Task memory & Attribution to retrieved source tasks. \\
Prediction confidence & Sample & Reliability of individual predictions. \\
\bottomrule
\end{tabular}%
}
\vspace{-0.4em} 
\end{table}

\begin{figure}[!htb]
\centering
\includegraphics[width=0.82\linewidth]{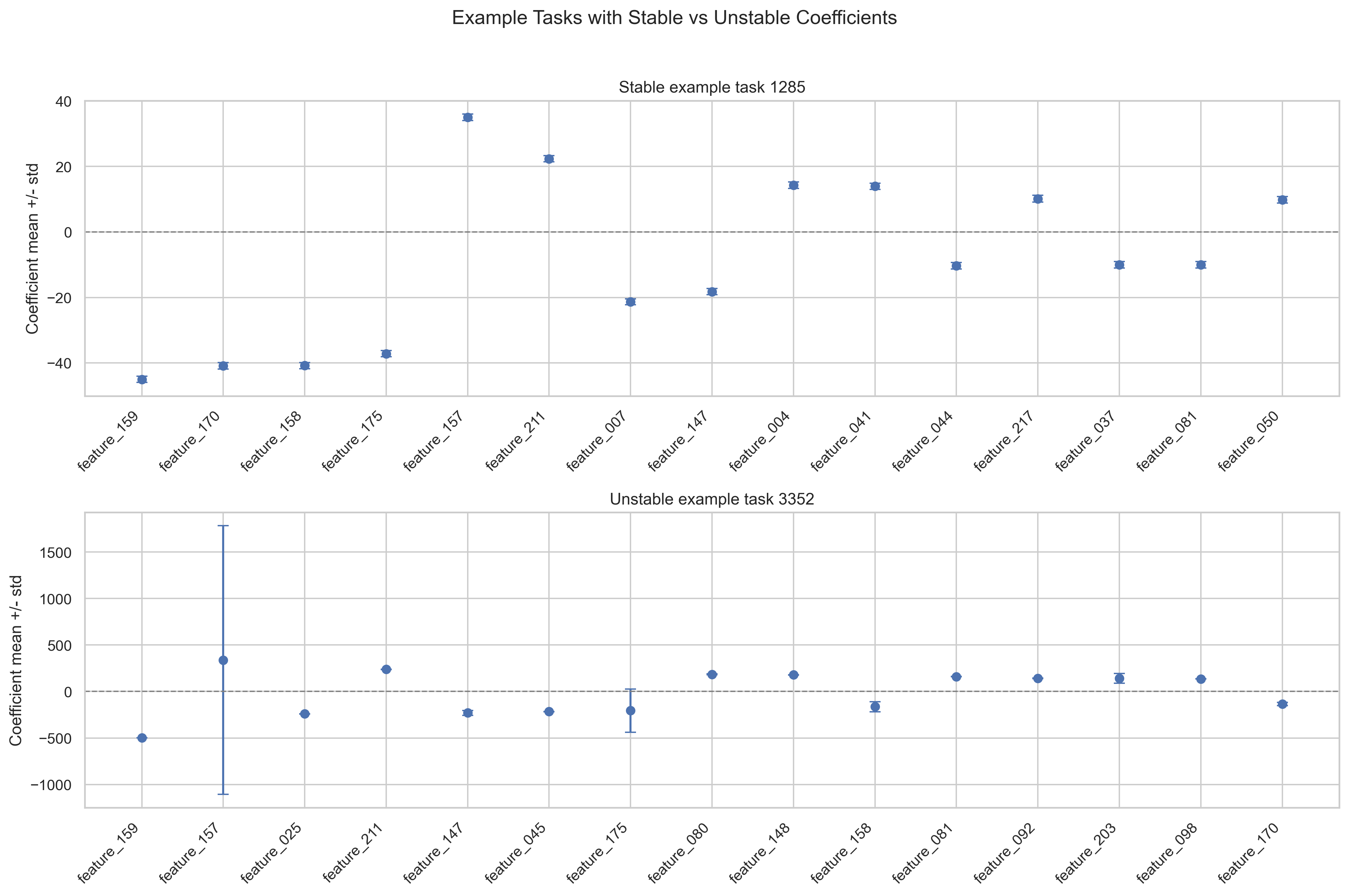}
\caption{Stable and unstable coefficient examples. Stable tasks exhibit top coefficients with relatively narrow posterior uncertainty intervals, whereas unstable tasks show wider intervals for some features and should therefore be interpreted with greater caution. 
}
\label{fig:app-stable-unstable-coefs}
\vspace{-0.6em} 
\end{figure}

{Tab.}~\ref{tab:runtime-scaling} summarizes the measured computational costs. Retrieval remains approximately constant as the memory grows from $100$ to $4{,}412$ tasks in the reported implementation, while total inference cost grows with the feature dimension.
An approximate-nearest-neighbor index can reduce the retrieval component for substantially larger memories.

\begin{table}[!ht]
\centering
\caption{Measured computational scaling and resource use. {Timings are reported per component.}}
\label{tab:runtime-scaling}
\begin{tabular}{lll}
\toprule
Quantity & Setting & Measured cost \\
\midrule
Retrieval & Memory size $100\rightarrow4{,}412$ & $0.13\rightarrow0.14$ ms \\
Coefficient generation & Candidate count $S=10\rightarrow200$ & Approximately $2.2$ ms \\
Inference & {Coefficient dimension $218\rightarrow869$ (RAIL$_{\mathrm{LR}}\rightarrow$RAIL$_{\mathrm{GAM}}$)} & $6.3\rightarrow{7.4}$ ms \\
Parametric memory & Full memory & $22$ MB \\
Training & RAIL training & $19.6$ s per epoch \\
\bottomrule
\end{tabular}
\vspace{-0.4em} 
\end{table}

\subsection{LLM-Based Task-Description Expansion}
\label{app:description-expansion}

To evaluate sensitivity to task-description quality, we expanded each procedure's original title into a longer, more detailed description using a large language model. We then re-embedded the expanded descriptions and retrained and evaluated RAIL with them.

\paragraph{Model and decoding.}
We used OpenAI's \texttt{gpt-4.1-mini} {(April 2025 release)}\footnote{{\url{https://openai.com/index/gpt-4-1/}}} through the Responses API with temperature $0.2$.
{Only the ICD-10-PCS procedure titles were expanded. Each request contained only the public procedure title (see the user prompt below); no patient-level MIMIC-IV data were sent to the API. All results outside this sensitivity analysis use the original titles.} Each unique procedure title was expanded exactly once, and the same expansion was reused for every occurrence of that title. Transient API failures, such as rate limits or timeouts, were retried up to six times with exponential backoff.

\paragraph{System prompt.}
The following system prompt was used for all requests:
\begin{quote}
\small\ttfamily
You are a clinical description editor. Expand brief procedure names into clear, factual, patient-friendly descriptions. Keep the meaning, do not invent codes, devices, or claims. Avoid brand names and prices. Aim for \textasciitilde{}120 words. Do not add contraindications, outcomes, or risks unless implied by the title. Keep it neutral and descriptive.
\end{quote}

\paragraph{User prompt.}
Each request contained a single user message built from the procedure's original long title:
\begin{quote}
\small\ttfamily
Procedure: "\{title\}"
\end{quote}
Here, \texttt{\{title\}} is replaced by the procedure's original long title.

\paragraph{Design rationale.}
The system prompt is designed to keep the expansion faithful to the original title. It forbids invented codes, devices, and claims, and it excludes outcomes, risks, and contraindications that the title does not imply. This is intended to provide richer phrasing of the same clinical content rather than new information about the target task.
\section{Additional Theoretical Analysis}
\label{app:additional-theory}

\subsection{ELBO derivation for learned retrieval}
\label{app:elbo-proof}

{We derive the discrete-latent ELBO underlying Eq.~\ref{eq:neg-elbo}; the soft-weighted implementation is specified separately below.} Recall the conditional generative model
\begin{align}
p_\phi(\mathcal D_t,\theta_t,I_t\mid c_t,\mathcal V)
=
p(\mathcal D_t\mid \theta_t)\;
p_\phi(\theta_t\mid I_t,c_t)\;
p(I_t\mid c_t,\mathcal V).
\label{eq:app-joint}
\end{align}
{Here, \(I_t\in\{1,\ldots,S\}\) indexes an entry in the candidate set \(\mathcal C_t\) selected using cosine-similarity scores. Both the prior and learned categorical retrieval distributions are defined over these \(S\) candidates. In implementation, the learned probabilities weight candidate contributions in the coefficient generator instead of sampling a discrete index.}

The marginal evidence for task \(t\) is
\begin{align}
p_\phi(\mathcal D_t\mid c_t,\mathcal V)
=
\sum_{I_t}
\int
p_\phi(\mathcal D_t,\theta_t,I_t\mid c_t,\mathcal V)
\,d\theta_t ,
\label{eq:app-evidence}
\end{align}
where the summation is over the support of the retrieval distribution.

For any variational distribution
\(q_\psi(\theta_t,I_t\mid \mathcal D_t,c_t,\mathcal V)\), we can write
\begin{align}
\log p_\phi(\mathcal D_t\mid c_t,\mathcal V)
&=
\log
\sum_{I_t}\int
q_\psi(\theta_t,I_t\mid \mathcal D_t,c_t,\mathcal V)
\frac{
p_\phi(\mathcal D_t,\theta_t,I_t\mid c_t,\mathcal V)
}{
q_\psi(\theta_t,I_t\mid \mathcal D_t,c_t,\mathcal V)
}
\,d\theta_t \\
&\ge
\mathbb E_{q_\psi(\theta_t,I_t\mid \mathcal D_t,c_t,\mathcal V)}
\left[
\log
\frac{
p_\phi(\mathcal D_t,\theta_t,I_t\mid c_t,\mathcal V)
}{
q_\psi(\theta_t,I_t\mid \mathcal D_t,c_t,\mathcal V)
}
\right],
\label{eq:app-jensen}
\end{align}
where the inequality follows from Jensen's inequality. Thus,
\begin{align}
\mathcal L_{\mathrm{ELBO}}^{(t)}
=
\mathbb E_{q_\psi}
\left[
\log p_\phi(\mathcal D_t,\theta_t,I_t\mid c_t,\mathcal V)
-
\log q_\psi(\theta_t,I_t\mid \mathcal D_t,c_t,\mathcal V)
\right].
\label{eq:app-elbo-def}
\end{align}

We use the restricted amortized factorization
\begin{align}
q_\psi(\theta_t,I_t\mid \mathcal D_t,c_t,\mathcal V)
=
q_\psi(I_t\mid c_t,\mathcal V)\;
q_\psi(\theta_t\mid \mathcal D_t,I_t,c_t).
\label{eq:app-q-factorization}
\end{align}
The retrieval factor is intentionally label-free. This restriction preserves zero-shot retrieval at inference time, where \(\mathcal D_t\) is unavailable. It is still a valid variational family, but it may yield a looser bound than a fully label-conditioned retrieval posterior.

Substituting Eq.~\ref{eq:app-joint} and Eq.~\ref{eq:app-q-factorization} into Eq.~\ref{eq:app-elbo-def} gives
\begin{equation}
\begin{aligned}
\mathcal L_{\mathrm{ELBO}}^{(t)}
&=
\mathbb E_{q_\psi}
\Big[
\log p(\mathcal D_t\mid\theta_t)
+
\log p_\phi(\theta_t\mid I_t,c_t)
+
\log p(I_t\mid c_t,\mathcal V)
\\
&\qquad\qquad
-
\log q_\psi(\theta_t\mid \mathcal D_t,I_t,c_t)
-
\log q_\psi(I_t\mid c_t,\mathcal V)
\Big].
\end{aligned}
\label{eq:app-elbo-expanded-logs}
\end{equation}
Grouping the likelihood, coefficient, and retrieval terms yields
\begin{equation}
\begin{aligned}
\mathcal L_{\mathrm{ELBO}}^{(t)}
&=
\mathbb E_{q_\psi(I_t\mid c_t,\mathcal V)}
\mathbb E_{q_\psi(\theta_t\mid \mathcal D_t,I_t,c_t)}
\big[
\log p(\mathcal D_t\mid\theta_t)
\big]
\\
&\quad+
\mathbb E_{q_\psi(I_t\mid c_t,\mathcal V)}
\mathbb E_{q_\psi(\theta_t\mid \mathcal D_t,I_t,c_t)}
\left[
\log
\frac{
p_\phi(\theta_t\mid I_t,c_t)
}{
q_\psi(\theta_t\mid \mathcal D_t,I_t,c_t)
}
\right]
\\
&\quad+
\mathbb E_{q_\psi(I_t\mid c_t,\mathcal V)}
\left[
\log
\frac{
p(I_t\mid c_t,\mathcal V)
}{
q_\psi(I_t\mid c_t,\mathcal V)
}
\right].
\end{aligned}
\label{eq:app-elbo-grouped}
\end{equation}
Recognizing the last two terms as negative KL divergences gives
\begin{equation}
\begin{aligned}
\mathcal L_{\mathrm{ELBO}}^{(t)}
&=
\mathbb E_{q_\psi(I_t\mid c_t,\mathcal V)}
\mathbb E_{q_\psi(\theta_t\mid \mathcal D_t,I_t,c_t)}
\big[
\log p(\mathcal D_t\mid\theta_t)
\big]
\\
&\quad-
\mathbb E_{q_\psi(I_t\mid c_t,\mathcal V)}
\Big[
\mathrm{KL}\!\left(
q_\psi(\theta_t\mid \mathcal D_t,I_t,c_t)
\,\middle\|\,
p_\phi(\theta_t\mid I_t,c_t)
\right)
\Big]
\\
&\quad-
\mathrm{KL}\!\left(
q_\psi(I_t\mid c_t,\mathcal V)
\,\middle\|\,
p(I_t\mid c_t,\mathcal V)
\right).
\end{aligned}
\label{eq:app-elbo-final}
\end{equation}
This leads to the negative ELBO objective in Eq.~\ref{eq:neg-elbo}. The first term trains the generated coefficients to explain the target-task labels. The second term anchors the coefficient posterior to the retrieval-conditioned coefficient prior. The third term trains the learned retriever relative to the cosine-similarity retrieval prior.

\paragraph{Closed-form KL terms.}
When both the coefficient posterior and prior are diagonal Gaussians,
\begin{align}
q_\psi(\theta_t\mid\cdot)
=
\mathcal N(\theta_t;\mu_{q,t},\mathrm{diag}(\sigma^2_{q,t})),
\qquad
p_\phi(\theta_t\mid\cdot)
=
\mathcal N(\theta_t;\mu_{\phi,t},\mathrm{diag}(\sigma^2_{\phi,t})),
\end{align}
the general diagonal Gaussian KL is
\begin{align}
\mathrm{KL}\big(q_\psi(\theta_t\mid\cdot)\,\|\,p_\phi(\theta_t\mid\cdot)\big)
=
\frac{1}{2}
\sum_{r=1}^d
\left[
\log\frac{\sigma_{\phi,t}^{2(r)}}{\sigma_{q,t}^{2(r)}}
+
\frac{\sigma_{q,t}^{2(r)}+(\mu_{q,t}^{(r)}-\mu_{\phi,t}^{(r)})^2}
{\sigma_{\phi,t}^{2(r)}}
-
1
\right].
\label{eq:kl-theta-general}
\end{align}
When the posterior and prior variances are tied,
\(\sigma^2_{q,t}\equiv\sigma^2_{\phi,t}\), Eq.~\ref{eq:kl-theta-general} simplifies to
\begin{align}
\mathrm{KL}\big(q_\psi(\theta_t\mid\cdot)\,\|\,p_\phi(\theta_t\mid\cdot)\big)
=
\frac{1}{2}\sum_{r=1}^d
\frac{(\mu_{q,t}^{(r)}-\mu_{\phi,t}^{(r)})^2}
{\sigma_{\phi,t}^{2(r)}}.
\label{eq:kl-theta}
\end{align}

{For the categorical retrieval distributions over the same \(S\) candidates,}
\(q_\psi(I_t\mid\cdot)=\mathrm{Cat}(\pi_t^{\mathrm{post}})\) and
\(p(I_t\mid\cdot)=\mathrm{Cat}(\pi_t^{\mathrm{prior}})\), the retrieval KL is
\begin{align}
\mathrm{KL}\big(q_\psi(I_t\mid\cdot)\,\|\,p(I_t\mid\cdot)\big)
=
\sum_{j=1}^S
\pi^{\mathrm{post}}_{t,j}
\log
\frac{\pi^{\mathrm{post}}_{t,j}}
{\pi^{\mathrm{prior}}_{t,j}}.
\label{eq:kl-I}
\end{align}

\paragraph{Practical estimator.}
{In implementation, the learned retrieval weights attenuate less relevant candidates inside the coefficient generator, yielding soft-weighted Gaussian prior and posterior parameters without sampling \(I_t\). We use one reparameterized coefficient sample \(\theta_t\) to estimate the likelihood term and compute the coefficient KL between these Gaussian distributions. The practical weighted surrogate objective is:}
\begin{align}
\widehat{\mathcal J}_t
=
\underbrace{{\textstyle\frac{1}{|\mathcal D_t|}\sum_{(x,y)\in\mathcal D_t}}\mathrm{BCE}\big(y,\sigma(\theta_t^\top {\Phi(x)})\big)}_{\text{data term}}
+
\beta_\theta
\underbrace{\mathrm{KL}_\theta}_{Eq.~\ref{eq:kl-theta}}
+
\beta_I
\underbrace{\mathrm{KL}_I}_{Eq.~\ref{eq:kl-I}}.
\label{eq:loss-practical}
\end{align}
{The data term is averaged over $\mathcal D_t$; the resulting rescaling relative to the KL terms is absorbed by the weights.} The weights \(\beta_\theta\) and \(\beta_I\) control the strength of coefficient-prior anchoring and retrieval-prior anchoring, respectively. {Even when \(\beta_\theta=\beta_I=1\), the use of soft retrieval weights makes this a differentiable surrogate, not an exact evaluation of the discrete-index negative ELBO.}

\subsection{Prior anchoring and generalization}
\label{app:prior-anchoring}
We provide a simple analysis based on standard Rademacher-complexity bounds \citep{bartlett2002rademacher,shalev2014understanding,mohri2018foundations}, showing why anchoring the coefficient posterior near a retrieval-conditioned prior can reduce the effective hypothesis class in low-data regimes. 
{It formalizes the intuition that, once the retrieved prior mean is fixed, constraining coefficients to remain near that mean controls the complexity of the induced linear predictors; the explicit norm constraint below serves as an analytical proxy for prior anchoring.}

\begin{assumption}[Bounded features and bounded Lipschitz loss]
\label{assump:features}
The {feature-mapped inputs} satisfy $\|{\Phi(x)}\|_2\le R$ almost surely. The loss
$\ell(\hat y,y)$ is $L_\ell$-Lipschitz in the prediction $\hat y$ and takes values in an interval of length at most $C_\ell$ over the hypothesis class considered below.
\end{assumption}

\begin{proposition}[{Generalization bound under prior anchoring}]
\label{prop:gen-bound}
Fix a retrieval-conditioned prior mean $\mu_{\phi,t}$ independently of the target-task sample used in the bound, and consider the hypothesis class
\begin{align}
\mathcal H_B
=
\{x\mapsto \theta^\top {\Phi(x)}:\|\theta-\mu_{\phi,t}\|_2\le B\}.
\end{align}
{Under Assumption~\ref{assump:features}, the} empirical Rademacher complexity satisfies
\begin{align}
\mathfrak R_m(\mathcal H_B)
\le
\frac{RB}{\sqrt m}.
\end{align}
Moreover, {with probability at least} $1-\delta$ over {an i.i.d. target-task sample} of size $m$, for all $\theta$ such that $\|\theta-\mu_{\phi,t}\|_2\le B$,
\begin{align}
\mathcal L(\theta)
\le
\widehat{\mathcal L}(\theta)
+
2L_\ell\,\mathfrak R_m(\mathcal H_B)
+
3C_\ell\sqrt{\frac{\log(2/\delta)}{2m}}
\le
\widehat{\mathcal L}(\theta)
+
\frac{2L_\ell RB}{\sqrt m}
+
3C_\ell\sqrt{\frac{\log(2/\delta)}{2m}}.
\end{align}
\end{proposition}

\paragraph{Proof.}
Let $S_m=\{x_1,\ldots,x_m\}$ be a fixed sample and let
$\sigma_1,\ldots,\sigma_m$ be independent Rademacher variables. The empirical Rademacher complexity of $\mathcal H_B$ is
\begin{align}
\mathfrak R_m(\mathcal H_B)
=
\mathbb E_\sigma
\left[
\sup_{\|\theta-\mu_{\phi,t}\|_2\le B}
\frac{1}{m}
\sum_{i=1}^m
\sigma_i \theta^\top {\Phi(x_i)}
\right].
\end{align}
Write $\theta=\mu_{\phi,t}+u$, where $\|u\|_2\le B$. Then
\begin{align}
\mathfrak R_m(\mathcal H_B)
&=
\mathbb E_\sigma
\left[
\sup_{\|u\|_2\le B}
\frac{1}{m}
\sum_{i=1}^m
\sigma_i(\mu_{\phi,t}+u)^\top {\Phi(x_i)}
\right] \\
&=
\mathbb E_\sigma
\left[
\frac{1}{m}
\sum_{i=1}^m
\sigma_i\mu_{\phi,t}^\top {\Phi(x_i)}
+
\sup_{\|u\|_2\le B}
u^\top
\left(
\frac{1}{m}
\sum_{i=1}^m
\sigma_i {\Phi(x_i)}
\right)
\right].
\end{align}
The first term vanishes in expectation over the Rademacher variables because
$\mathbb E_\sigma[\sigma_i]=0$. Therefore,
\begin{align}
\mathfrak R_m(\mathcal H_B)
=
\mathbb E_\sigma
\left[
\sup_{\|u\|_2\le B}
u^\top
\left(
\frac{1}{m}
\sum_{i=1}^m
\sigma_i {\Phi(x_i)}
\right)
\right].
\end{align}
By Cauchy--Schwarz,
\begin{align}
\sup_{\|u\|_2\le B}
u^\top
\left(
\frac{1}{m}
\sum_{i=1}^m
\sigma_i {\Phi(x_i)}
\right)
\le
\frac{B}{m}
\left\|
\sum_{i=1}^m
\sigma_i {\Phi(x_i)}
\right\|_2 .
\end{align}
Thus,
\begin{align}
\mathfrak R_m(\mathcal H_B)
\le
\frac{B}{m}
\mathbb E_\sigma
\left\|
\sum_{i=1}^m
\sigma_i {\Phi(x_i)}
\right\|_2 .
\end{align}
Applying Jensen's inequality,
\begin{align}
\mathbb E_\sigma
\left\|
\sum_{i=1}^m
\sigma_i {\Phi(x_i)}
\right\|_2
&\le
\left(
\mathbb E_\sigma
\left\|
\sum_{i=1}^m
\sigma_i {\Phi(x_i)}
\right\|_2^2
\right)^{1/2}.
\end{align}
Expanding the squared norm gives
\begin{align}
\mathbb E_\sigma
\left\|
\sum_{i=1}^m
\sigma_i {\Phi(x_i)}
\right\|_2^2
&=
\mathbb E_\sigma
\left[
\sum_{i=1}^m\sum_{j=1}^m
\sigma_i\sigma_j {\Phi(x_i)}^\top{\Phi(x_j)}
\right] \\
&=
\sum_{i=1}^m
\|{\Phi(x_i)}\|_2^2,
\end{align}
because $\mathbb E[\sigma_i\sigma_j]=0$ for $i\neq j$ and
$\mathbb E[\sigma_i^2]=1$. Since $\|{\Phi(x_i)}\|_2\le R$,
\begin{align}
\sum_{i=1}^m\|{\Phi(x_i)}\|_2^2
\le
mR^2.
\end{align}
Therefore,
\begin{align}
\mathfrak R_m(\mathcal H_B)
\le
\frac{B}{m}\sqrt{mR^2}
=
\frac{RB}{\sqrt m}.
\end{align}

It remains to connect this complexity bound to generalization. By the standard Rademacher generalization theorem for a loss class with range at most $C_\ell$, with probability at least $1-\delta$, uniformly for all $h\in\mathcal H_B$,
\begin{align}
\mathcal L(h)
\le
\widehat{\mathcal L}(h)
+
2\mathfrak R_m(\ell\circ\mathcal H_B)
+
3C_\ell\sqrt{\frac{\log(2/\delta)}{2m}}.
\end{align}
Since $\ell$ is $L_\ell$-Lipschitz in the prediction, the contraction inequality gives
\begin{align}
\mathfrak R_m(\ell\circ\mathcal H_B)
\le
L_\ell \mathfrak R_m(\mathcal H_B).
\end{align}
Substituting the Rademacher-complexity bound derived above yields
\begin{align}
\mathcal L(\theta)
\le
\widehat{\mathcal L}(\theta)
+
2L_\ell\mathfrak R_m(\mathcal H_B)
+
3C_\ell\sqrt{\frac{\log(2/\delta)}{2m}}
\le
\widehat{\mathcal L}(\theta)
+
\frac{2L_\ell RB}{\sqrt m}
+
3C_\ell\sqrt{\frac{\log(2/\delta)}{2m}}.
\end{align}
This completes the proof. \qed


\end{document}